\documentclass[letterpaper]{article} 
\usepackage{aaai2027}  
\usepackage[hyphens]{url}  
\usepackage{graphicx} 
\usepackage{natbib}  
\usepackage{caption} 
\usepackage{algorithm}
\usepackage{algorithmic}

\usepackage{amsmath}
\usepackage{amssymb}
\usepackage{nicematrix} 

\usepackage{newfloat}
\usepackage{listings}
\DeclareCaptionStyle{ruled}{labelfont=normalfont,labelsep=colon,strut=off} 
\floatstyle{ruled}
\newfloat{listing}{tb}{lst}{}
\floatname{listing}{Listing}

\usepackage{booktabs}

\title{ZENITH: An Algorithm for Automated Learning Rate Adaptation}

\author{
    Dhrubo Saha
}

\affiliations{
    Georgia Institute of Technology\\
    dsaha36@gatech.edu
}

\usepackage{multirow}       
\usepackage{tabularx}       
\usepackage{makecell}

\newcolumntype{Y}{>{\centering\arraybackslash}X}

\definecolor{lightgray}{gray}{0.93}
\definecolor{bestgreen}{rgb}{0.0, 0.5, 0.0}

\newcommand{\B}[1]{\textbf{#1}}

\nocopyright
\begin{document}
\maketitle

\begin{abstract}
Training deep learning models requires manual oversight or tuning of the learning rate (LR) schedule. While existing adaptive schedulers dynamically adjust the LR, they suffer from high computational and memory overhead, incompatibility with regularization, and suboptimal LR trajectories. To address these limitations, we introduce the ZENITH scheduler, which adapts the LR using the temporal evolution of gradient signals. Theoretical and empirical analyses show that it is guaranteed to converge, and is inclined toward flatter minima with better generalization. Across six CNN architectures and six benchmarks for image classification, the R-CNN family for detection and segmentation, and MLPs for tabular regression, experiments demonstrate that ZENITH achieves better model performance in less wall-clock time than existing methods.
\end{abstract}


\section{Introduction}
Deep learning models are trained via gradient descent, which requires an appropriate initial learning rate (LR) and this LR's manual scheduling. These hyperparameter choices influence both the model performance and the wall-clock training time required. This challenge is exacerbated by the current trend of training large-scale models on massive datasets, where longer training durations make manual LR scheduling more labor-intensive. 

To avoid manual oversight, practitioners employ predefined LR schedules like exponential decay \cite{szegedy2016rethinking}, step decay \cite{ge2019step}, and cosine annealing \cite{loshchilov2016sgdr}. These schedules adapt the LR via fixed intervals or continuous functions. However, they are often impractical because the optimal number of iterations and decay rate are not known in advance. Consequently, tuning these parameters becomes a costly trial-and-error process for every new task. To address this challenge, researchers have been developing automatic, parameter-free schedulers.

\subsection{Related Work}
\textbf{Coin Betting.} One of the first automatic schedulers was COCOB, which treats weight updates as coin bets and derives the LR from the accumulated reward from previous iterations \cite{orabona2017training}. If the gradients point in a consistent direction, the LR increases to accelerate convergence, and vice versa. However, the noisy gradients of mini-batch training cause the algorithm to underestimate the ideal LR. Moreover, COCOB needs 6 times as much memory as vanilla SGD because it also has to store each weight's maximum observed gradient, sum of absolute gradients, accumulated reward, sum of gradients, and initial betting states.

\textbf{Quadratic Loss Approximation.} Quadratic approximation methods like PAL \cite{mutschler2020parabolic} and QLABGrad \cite{fu2024qlabgrad} assume that the local loss landscape can be modeled as a quadratic function. In each iteration, the algorithm computes the loss and gradients at the current weights. It then takes a probing step in the descent direction and evaluates the loss at this second point. Next, it derives the equation for the parabola to locate its vertex, reaching it in a single step. However, this process introduces a new hyperparameter, the probing step size, which may need to be tuned in lieu of the LR schedule. Furthermore, the mechanism needs 1 extra forward pass per iteration (or even 2 in curvature-estimating extensions \cite{zhu2021automatic}), inflating the wall-clock training time. There is also a substantial memory footprint for storing additional copies of the weights. Although the GeN algorithm \cite{bu2024gradient} reduced these burdens by performing approximations only every 4--8 iterations, this compromises the method's efficacy.

\textbf{Distance-Aware Estimation.} DoG \cite{ivgi2023dog} and DoWG \cite{khaled2023dowg} derive the LR by normalizing the distance traveled by the accumulated gradient norms. Similarly, D-Adaptation \cite{defazio2023learning} and Prodigy \cite{mishchenko2023prodigy} estimate the distance to the solution through lower-bound estimations. Unlike normal training, which allows each weight update to be computed independently, these schedulers depend on aggregate statistics that must be calculated globally, increasing the per-iteration wall-clock time. Additionally, they are sensitive to their LR growth rate. Setting this parameter too low results in excessively small step sizes, and vice versa. Moreover, regularization can destabilize their LR estimates by making the solution appear farther away. Lastly, they are memory-intensive. For instance, D-Adaptation requires 4 times the memory of vanilla SGD, as it must store the current weights, initial weights, and the $s$ and $z$ buffers.

\textbf{Polyak-Style Interpolation.} The Polyak-style methods, ALIG \cite{berrada2020training} and SPS \cite{loizou2021stochastic}, set the LR ($\eta_t$) equal to the ratio of the scalar loss to the squared gradient norm:
\begin{equation}
    \eta_t = \min \left( \frac{\mathcal{L}(\theta_t)}{\| \nabla\mathcal{L}(\theta_t) \|^2}, \eta_{\max} \right)
\end{equation}
The loss term decays the LR from a high initial value during early training to near-zero as the loss approaches zero. This approach is simple, but it yields better model performance and convergence speed than previous methods. However, it faces three key issues:

\textbf{Issue 1: }It assumes that convergence happens at a near-zero loss, which does not always hold in practice. For example, advanced detection and segmentation tasks often converge at losses far above zero. As a result, the method can substantially overestimate $\eta_t$, sometimes even defaulting to $\eta_{\max}$ throughout training.

\textbf{Issue 2: }It is sensitive to the absolute scales of the training loss and gradient norm. Factors like the loss function, regularization magnitude, and architectural complexity affect the ratio of these two quantities greatly. Consequently, the LR may become too small, causing sluggish convergence, or too large, constantly hitting its clipping bound. Although the L4 algorithm \cite{rolinek2018l4} tried to mitigate this instability by scaling the ratio with a factor $\alpha$, tuning $\alpha$ can be as costly as tuning the LR schedule.

\textbf{Issue 3: }It makes the LR inversely related to the gradient norm, arguing that steeper gradients warrant a smaller LR for stability. However, their LR dynamics are driven primarily by the loss, and the norm often causes LR to be underestimated.

\subsection{Our Contributions}

We propose ZENITH (\textbf{Z}ero-overhead \textbf{E}volution using \textbf{N}orm-\textbf{I}nformed \textbf{T}raining \textbf{H}istory), a novel scheduler for automatic LR adaptation. ZENITH introduces a principle fundamentally different from that of Polyak-style schedulers. Instead of tying the LR to the loss, it ties it to the temporal evolution of gradient norms. This results in a scheduling mechanism that is stable, scale-invariant, and broadly applicable. Our method addresses the aforementioned issues of Polyak-style schedulers through three design choices:

\textbf{(1) Gradient-driven LR dynamics:} The gradient norm provides an indication of our position in the loss landscape. This quantity is high during early training and diminishes toward zero near a minimum point. Therefore, we propose a positive relationship between the LR and the gradient norm. This maintains a high early LR to find flatter global minima. As the gradients attenuate, the LR drops to converge stably. Unlike the training loss, the gradient norm is predictably high at the beginning and low near a minimum.

\textbf{(2) Scale-invariant normalization.} Instead of using raw gradient norms, we normalize them using their historical smoothed maximum. This removes dependence on absolute scale, making the method robust across different models, datasets, and loss functions.

\textbf{(3) Proportional scaling for generalization.} Polyak-style schedulers make the LR inversely proportional to the gradient magnitude. However, through theoretical and empirical analyses, we show that this is suboptimal for LR scheduling because of its effect on minima flatness and generalization. Therefore, we propose a proportional relationship instead.

\section{Methodology}

\subsection{Algorithm Description}
ZENITH maintains a sliding window of the $W$ most recent gradient $L_2$ norms to compute a smoothed local estimate of the loss landscape’s steepness. Let $g_t$ denote the gradient norm at training iteration $t$. The algorithm maintains a First-In-First-Out queue, denoted as $Q$, with a fixed capacity $W$. At each iteration, the current norm $g_t$ is appended to $Q$, and if the size of $Q$ exceeds $W$, the oldest element is removed. Once the window is fully populated, we compute the rolling mean, $\mu_t$, of the values in $Q$:
\begin{equation}
    \mu_t = \frac{1}{W} \sum_{k=0}^{W-1} Q[k]
\end{equation}
The algorithm tracks the historical maximum (or zenith) of this rolling mean, denoted by $Z_t$. It does not track the maximum $g_t$, making it insensitive to noise and outliers. $Z_t$ serves as a reference point for the highest (smoothed) steepness observed during training and is updated monotonically:
\begin{equation}
    Z_t = \max (Z_{t-1}, \mu_t)
\end{equation}
ZENITH uses the initial LR $\eta_0$ until $Q$ is full, after which $\eta_t$ is scaled at each iteration by the ratio of the current local steepness to the historical zenith. Therefore, $\eta_t$ anneals as $\mu_t$ attenuates relative to its peak magnitude $Z_t$. $\eta_t$ is given to the base optimizer as:
\begin{equation}\label{eq:zenith_lr}
    \eta_t = \eta_0 \cdot \left( \frac{\mu_t}{Z_t} \right)
\end{equation}
$\eta_0$ is required by both ZENITH and baselines because these methods are designed to automate LR decay for a given $\eta_0$. These schemes are not intended for choosing the initial LR. Therefore, we conduct experiments where all methods use the same $\eta_0$, which fairly evaluates their scheduling efficacy. In further experiments, we also perform grid searches over $\eta_0$ to study each method’s sensitivity and tuned performance. Apart from $\eta_0$, ZENITH's only parameter is the window size $W$. In this work, we fix $W=5000$ across the experiments to show that ZENITH achieves strong performance without tuning $W$. The pseudocode is detailed in Algorithm 1.

\begin{algorithm}[tb]
\caption{ZENITH Adaptive Learning Rate Schedule}
\label{alg:zenith}
\textbf{Input}: Initial learning rate $\eta_0$, Window size $W$\\
\textbf{Initialize}: Window queue $Q \leftarrow \emptyset$, Zenith $Z \leftarrow 0$, Current LR $\eta \leftarrow \eta_0$
\begin{algorithmic}
\FOR{each iteration $t$}
\STATE Compute gradient $L_2$ norm: $g \leftarrow \|\nabla \mathcal{L}(\theta_t)\|_2$
\STATE Push $g$ to $Q$
\IF{size($Q$) $> W$}
\STATE Pop oldest value from $Q$
\ENDIF
\IF{size($Q$) $= W$}
\STATE Compute rolling mean: $\mu \leftarrow \text{mean}(Q)$
\STATE Update Zenith: $Z \leftarrow \max(Z, \mu)$
\STATE \COMMENT{Apply Decay}
\IF{$Z > 0$}
\STATE $\eta \leftarrow \eta_0 \cdot (\mu / Z)$
\ELSE
\STATE $\eta \leftarrow \eta_0$
\ENDIF
\STATE Update optimizer learning rate to $\eta$
\ENDIF
\ENDFOR
\end{algorithmic}
\end{algorithm}

\subsection{Theoretical Convergence Analysis}

\textbf{Update Rule A.} The weights are updated via gradient descent:
\begin{equation} \label{eq:update_rule}
    \theta_{t+1} = \theta_t - \eta_t \nabla\mathcal{L}(\theta_t)
\end{equation}
\textbf{Proposition B.} The current local gradient-norm estimate $\mu_t$ never exceeds its historical smoothed maximum $Z_t$. For all iterations where $Z_t > 0$:
\begin{equation} \label{eq:prop_b}
    \frac{\mu_t}{Z_t} \le 1 \implies \eta_t \le \eta_0
\end{equation}
We adopt the assumptions for smooth non-convex optimization from prior work \cite{fu2024qlabgrad}, with further details available there. Some of these assumptions include:

\textbf{Assumption C.} The loss function $\mathcal{L}$ is bounded below by a scalar $\mathcal{L}^*$:
\begin{equation} \label{eq:assump_c}
    \mathcal{L}(\theta) \ge \mathcal{L}^*, \quad \forall \theta \in \mathbb{R}^d
\end{equation}

\textbf{Assumption D.} The gradient of the loss function is $M$-Lipschitz continuous:
\begin{equation} \label{eq:assump_d}
    \| \nabla\mathcal{L}(x) - \nabla\mathcal{L}(y) \| \le M \| x - y \|,
    \quad \forall x,y \in \mathbb{R}^d
\end{equation}

\textbf{Theorem 1.} Under Assumptions C and D, if \(\eta_0<2/M\), ZENITH converges to a stationary point:
\[
    \lim_{t\to\infty}\|\nabla\mathcal{L}(\theta_t)\|=0.
\]
\textbf{Proof.} Using the $M$-Lipschitz smoothness criterion \eqref{eq:assump_d}, we invoke the Descent Lemma inequality:
\begin{equation} \label{eq:descent_lemma}
    \mathcal{L}(\theta_{t+1}) \le \mathcal{L}(\theta_t) + \langle \nabla\mathcal{L}(\theta_t), \theta_{t+1} - \theta_t \rangle + \frac{M}{2} \| \theta_{t+1} - \theta_t \|^2
\end{equation}
Substituting the update rule \eqref{eq:update_rule} into inequality \eqref{eq:descent_lemma}:
\begin{equation} \label{eq:descent_sub}
    \mathcal{L}(\theta_{t+1}) \le \mathcal{L}(\theta_t) - \eta_t \| \nabla\mathcal{L}(\theta_t) \|^2 \left(1 - \frac{M}{2}\eta_t\right)
\end{equation}
For the loss to decrease (i.e., $\mathcal{L}(\theta_{t+1}) \le \mathcal{L}(\theta_t)$), the term in the parentheses in inequality \eqref{eq:descent_sub} must be non-negative:
\begin{equation} \label{eq:loss_decrease}
    1 - \frac{M}{2}\eta_t > 0 \implies \eta_t < \frac{2}{M}
\end{equation}
Because of the Boundedness Property \eqref{eq:prop_b}, condition \eqref{eq:loss_decrease} is satisfied for all iterations $t$ if
\begin{equation} \label{eq:eta0_bound}
    \eta_0 < \frac{2}{M}
\end{equation}
Therefore, the step size remains stable because $\eta_t$ is bounded above by $\eta_0$, and it respects the Lipschitz stability bound provided $\eta_0$ is set appropriately. Recall that the Descent Inequality \eqref{eq:descent_sub} included the term $(1 - M\eta_t/2)$. Using the Boundedness Property \eqref{eq:prop_b}, this term can be bounded from below by a constant:
\begin{equation} \label{eq:safety_margin}
    1 - \frac{M}{2}\eta_t \ge 1 - \frac{M}{2}\eta_0
\end{equation}
Let us define this guaranteed safety margin as $C = 1 - M\eta_0 / 2$. If $\eta_0 < 2/M$ as required in inequality \eqref{eq:eta0_bound}, then $C > 0$. Substituting this constant $C$ and the definition of $\eta_t$ \eqref{eq:zenith_lr} into the Descent Inequality \eqref{eq:descent_sub} yields:
\begin{equation} \label{eq:rearranged}
    \frac{\mu_t}{Z_t} \| \nabla\mathcal{L}(\theta_t) \|^2 \le \frac{\mathcal{L}(\theta_t) - \mathcal{L}(\theta_{t+1})}{C \eta_0}
\end{equation}
Summing the LHS of inequality \eqref{eq:rearranged} over $T$ iterations:
\begin{equation} \label{eq:summation}
    \sum_{t=W}^{T} \frac{\mu_t}{Z_t} \| \nabla\mathcal{L}(\theta_t) \|^2 \le \frac{\mathcal{L}(\theta_W) - \mathcal{L}(\theta_{T+1})}{C \eta_0} \le \frac{\mathcal{L}(\theta_W) - \mathcal{L}^*}{C \eta_0}
\end{equation}
The RHS of inequality \eqref{eq:summation} is finite. Therefore, the summation on the LHS must also be finite:
\begin{equation} \label{eq:inf_sum}
    \lim_{T \to \infty} \sum_{t=W}^{T} \frac{\mu_t}{Z_t} \| \nabla\mathcal{L}(\theta_t) \|^2 < \infty
\end{equation}
Inequality \eqref{eq:inf_sum} states that the infinite sum of non-negative terms is finite. By the properties of convergent series, the $t$-th term must approach zero as $t \to \infty$:
\begin{equation} \label{eq:zero_limit}
    \lim_{t \to \infty} \left( \frac{\mu_t}{Z_t} \| \nabla\mathcal{L}(\theta_t) \|^2 \right) = 0
\end{equation}
The rolling mean $\mu_t$ is bounded below by any single element in its window divided by $W$, meaning $\mu_t \ge \frac{1}{W} \| \nabla\mathcal{L}(\theta_t) \|$. Since $Z_t$ is a non-decreasing and bounded sequence, $Z_t \le Z_\infty$ for some finite positive constant $Z_\infty$. Substituting these bounds into condition \eqref{eq:zero_limit} yields:
\begin{equation} \label{eq:strict_bounds}
    \lim_{t \to \infty} \frac{1}{W Z_\infty} \| \nabla\mathcal{L}(\theta_t) \|^3 \le \lim_{t \to \infty} \left( \frac{\mu_t}{Z_t} \| \nabla\mathcal{L}(\theta_t) \|^2 \right) = 0
\end{equation}
Since $W$ and $Z_\infty$ are positive finite constants, $\nabla\mathcal{L}(\theta_t)$ must vanish, proving convergence to a stationary point. $\square$

\section{Experiments}

\subsection{Experimental Setup}

\begin{table}[b]
\centering
\renewcommand{\arraystretch}{1.15}

\begin{NiceTabularX}{\columnwidth}{c | X X c}
    \CodeBefore
        \rowcolors{2}{white}{lightgray}
    \Body
    \hline 
    
    \textbf{Exp} & \textbf{Dataset} & \textbf{Model} & \textbf{Time (hr)} \\
    \hline 
    
    A & MNIST & EfficientNet-B0 & 1 \\
    B & CIFAR-10 & VGG19 & 1 \\
    C & CIFAR-100 & ResNet50 & 1 \\
    \hline 
    D & Food-101 & DLA & 5 \\
    E & Tiny ImageNet & Inception & 5 \\
    \hline
    F & ImageNet-100 & DenseNet121 & 10 \\
    \hline 
\end{NiceTabularX}
\caption{Setup for image classification benchmarks.}
\label{tab:experiment_setup}
\end{table}

\begin{figure*}[t]
    \centering
    \includegraphics[width=\textwidth]{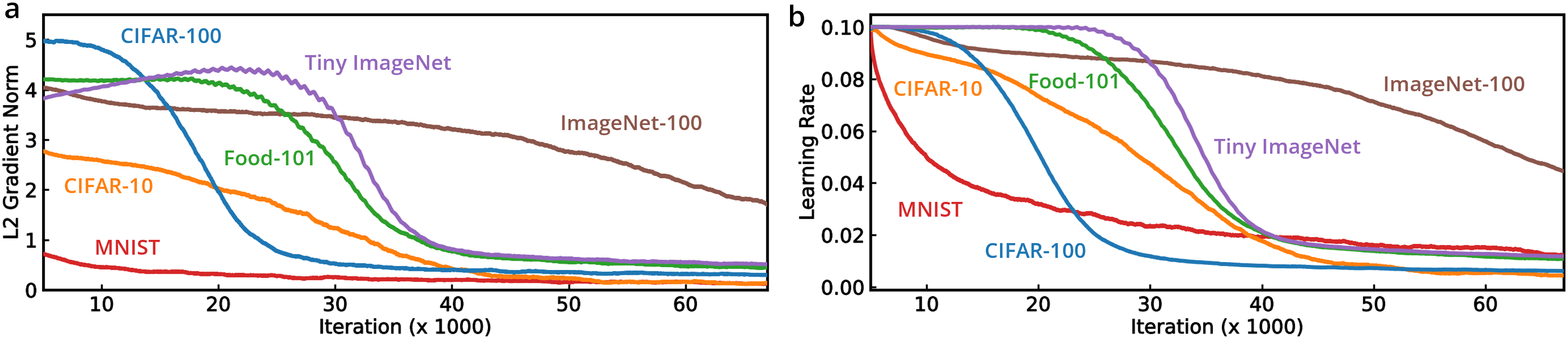} 
    \caption{(a) Evolution of the $L_2$ gradient norm and (b) the corresponding LR used by ZENITH across iterations.}
    \label{fig:lr_decay}
\end{figure*}

We conducted experiments across 6 combinations of datasets \cite{lecun2002gradient, krizhevsky2009learning, bossard2014food, le2015tiny, deng2009imagenet} and model architectures \cite{simonyan2014very, he2016deep, szegedy2016rethinking, huang2017densely, yu2018deep, tan2019efficientnet}, as detailed in Table~\ref{tab:experiment_setup}. Data augmentation was used, including random cropping, horizontal flipping, translations, and rotations. The cross-entropy loss function was utilized. Models were trained using a batch size of 128 on an NVIDIA A100 GPU with six data loading workers. Training durations were set for each dataset-architecture pair to ensure sufficient time for baselines to converge (see Table~\ref{tab:experiment_setup}). Test accuracy was evaluated at the end of each epoch for reporting, which is the standard procedure in this research area. Cumulative wall-clock time was recorded for the training phase exclusively, excluding testing time. $L_2$ regularization was not used because some baselines perform poorly with regularization. Three runs were conducted using random seeds 42, 43, and 44. The base optimizer was SGD, as many baselines are only applicable to SGD and SGD is commonly used for CNN-based image classification.

ZENITH was evaluated against the 12 baselines discussed in the Related Work section. All these methods were recently published in the NeurIPS, ICML, ICLR, AAAI, and AISTATS conferences, and the journal Neural Networks. This attests to their competitiveness and highlights the significance of automatic LR scheduling advancements in top AI venues. The initial and/or maximum LR was fixed at $0.1$ to ensure a fair comparison. This is the exact setup for fair comparisons used in recent AAAI'24 baselines \cite{fu2024qlabgrad}. For distance-aware estimators, we used an LR growth rate of 1.0, which was mandated in their GitHub repositories. The PyTorch and TensorFlow implementations of ZENITH, along with Python notebooks to reproduce the experiments, will be made available upon acceptance.

\begin{table}[!b]
\centering
\renewcommand{\arraystretch}{1.15}

\newcolumntype{Y}{>{\centering\arraybackslash}X}

\begin{NiceTabularX}{\columnwidth}{l | Y Y Y | Y}
    \CodeBefore
        \rowcolors{3}{lightgray}{white}
    \Body
    \hline 
    
    \Block{2-1}{\textbf{Method}} & \multicolumn{3}{c|}{\textbf{Time ($\times$SGD)}} & \textbf{Space} \\
    \cline{2-4} 
    
     & \textbf{Exp-B} & \textbf{Exp-D} & \textbf{Exp-F} & \textbf{($\times$Model)} \\
    \hline 
    
    SGD & 1.00 & 1.00 & 1.00 & 0 \\
    \hline
    
    PAL & 1.50 & 1.20 & 1.51 & 1 \\
    LQA & 4.72 & 2.33 & 3.72 & 5 \\
    QLAB & 1.20 & 1.14 & 1.38 & 1 \\
    GeN & 1.29 & 1.21 & 1.32 & 1 \\
    \hline

    COCOB & 1.50 & 1.06 & 1.34 & 5 \\
    \hline
    
    DoG & 1.16 & 1.01 & 1.10 & 1 \\
    DoWG & 1.28 & 1.05 & 1.21 & 1 \\
    DAdapt & 1.40 & 1.07 & 1.27 & 3 \\
    Prodigy & 1.69 & 1.14 & 1.44 & 4 \\
    \hline

    L4 & 1.83 & 1.18 & 1.50 & 2 \\
    ALIG & 1.18 & 1.03 & 1.10 & \underline{\B{0}} \\
    SPS & 1.18 & 1.06 & 1.17 & \underline{\B{0}} \\
    \hline
    
    \textbf{ZENITH} & \underline{\B{1.01}} & \underline{\B{1.01}} & \underline{\B{1.01}} & \underline{\B{0}} \\
    \hline 
\end{NiceTabularX}
\caption{Comparison of time and space overheads.}
\label{tab:overhead_comparison}
\end{table}

\subsection{Experimental Results and Discussion}

\begin{figure*}[t]
    \centering
    \includegraphics[width=0.96875\textwidth]{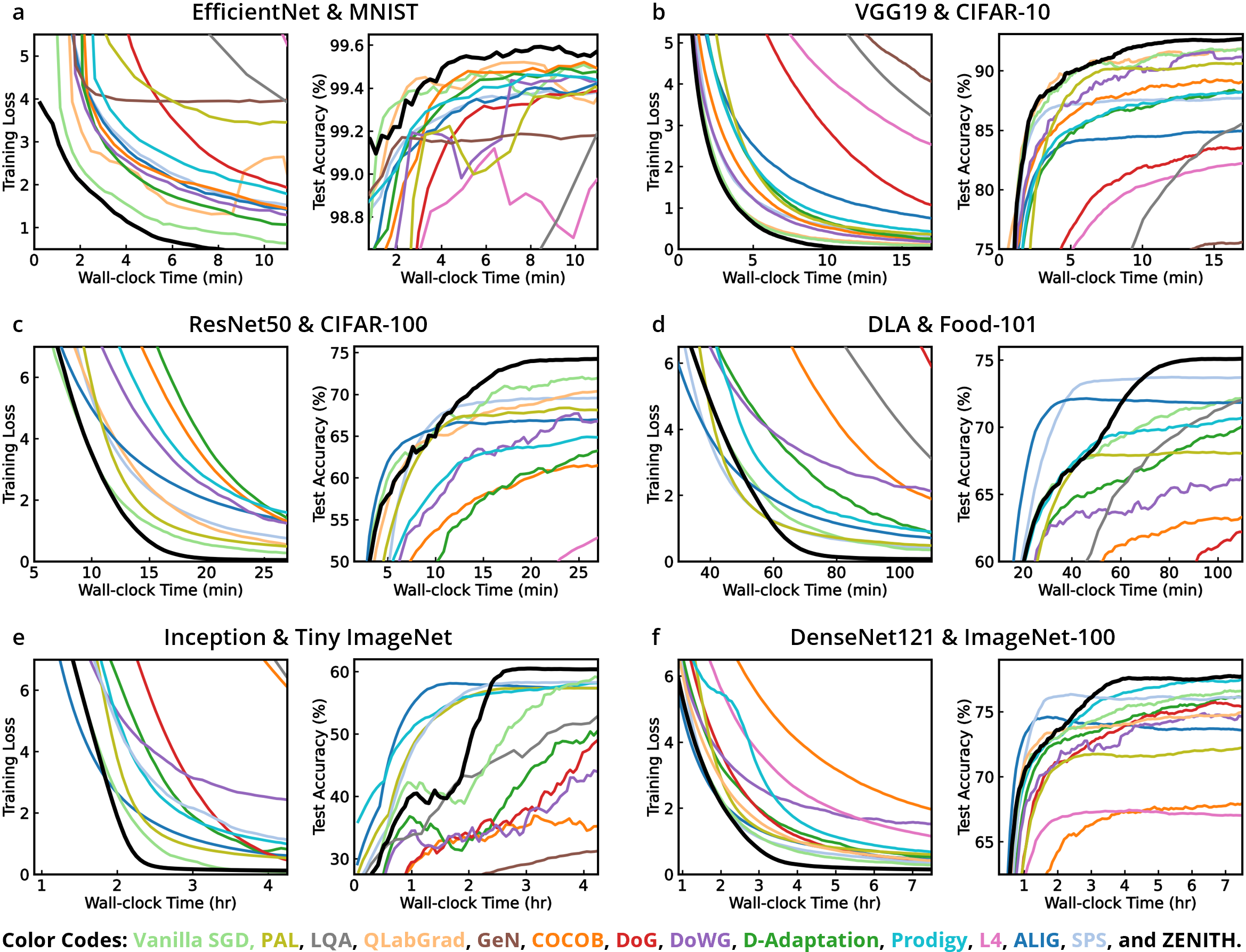} 
    \caption{Training loss and test accuracy curves against wall-clock time for the 6 image classification experiments. Training loss magnitudes in this figure are scaled by a factor of 10 or 100 for visual clarity. The loss function was not scaled during the actual training process. Additionally, all curves are smoothed using a simple moving average for visual clarity.}   
    \label{fig:image_classification}
\end{figure*}

\subsubsection{ZENITH's Scheduling Mechanism.}
Figure~\ref{fig:lr_decay} displays the evolution of the gradient norm and the corresponding LR adaptation of ZENITH. Figure~\ref{fig:lr_decay}a shows that the gradients may initially increase or plateau during the early phase of entering a loss basin. During this period, Figure~\ref{fig:lr_decay}b shows that ZENITH maintains a high LR. As we approach a minimum, the gradient norm diminishes, and ZENITH performs commensurate LR decay. Interestingly, the LR trajectories in Figure~\ref{fig:lr_decay}b mirror well-known manual schedules. The MNIST trajectory resembles an exponential decay, while the CIFAR-10 and ImageNet-100 trajectories resemble cosine annealing. The remaining three trajectories resemble a smoothed step decay. However, these manual schedules require prior knowledge of the optimal total iteration count or decay rate, both of which vary greatly across configurations. For example, while both CIFAR-10 and ImageNet-100 exhibit cosine annealing patterns in Figure~\ref{fig:lr_decay}b, ImageNet-100 needs over twice as many iterations. ZENITH resolves this issue by modulating the LR using gradient signals.

\subsubsection{Overhead Analysis.} Table~\ref{tab:overhead_comparison} compares the computational time and space overheads. To assess the time overhead, the wall-clock time per iteration was recorded across the experiments. ZENITH incurs only 1\% time penalty relative to vanilla SGD, whereas baselines increase iteration time by up to 372\%. This is because ZENITH's only significant operation beyond SGD is the calculation of the gradient norm. Furthermore, baselines require memory buffers of up to 5 times the model size. Conversely, ZENITH incurs zero model-sized memory overhead, as it only tracks fixed-size scalar statistics that are independent of model dimensions.

\subsubsection{Main Comparison.} ZENITH's adaptive scheduling in Figure~\ref{fig:lr_decay}, together with its faster iteration speed in Table~\ref{tab:overhead_comparison}, allows it to reach higher accuracies in less wall-clock time. Figure~\ref{fig:image_classification} compares the training-loss and test-accuracy curves of all algorithms as a function of wall-clock time. On MNIST, CIFAR-10, and CIFAR-100, ZENITH achieves higher test accuracy while requiring less wall-clock time than the baselines. While ALIG and SPS converge earlier on Food-101, Tiny-ImageNet, and ImageNet-100, this premature convergence compromises their test accuracy. No baseline across any experiment outperformed ZENITH on both objectives simultaneously. While baselines may perform well sporadically, ZENITH remains robust across experiments. ZENITH is the only automatic scheduler that achieves higher test accuracy than vanilla SGD, while requiring only 54\% of the training time. Tabulated results are in Appendix A, and all appendices are present in the supplementary material.

\begin{figure*}[t]
    \centering
    \includegraphics[width=\textwidth]{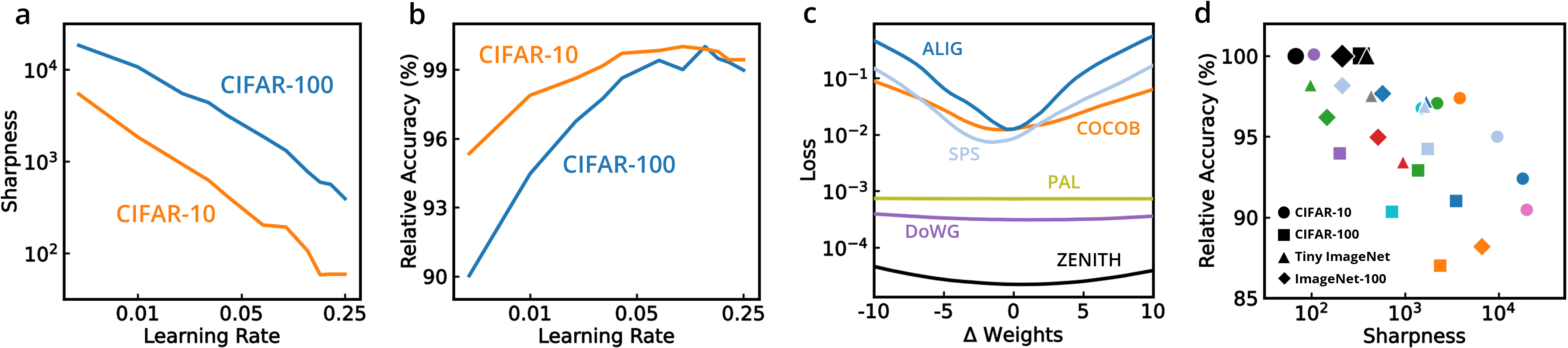} 
    \caption{Minima Sharpness and Generalization. (a-b) Influence of the LR on minima sharpness and test accuracy for vanilla SGD; (c) loss landscapes around the minima in the CIFAR-10 experiment; (d) accuracy versus sharpness across 4 experiments.}
    \label{fig:sharpness}
\end{figure*}

\begin{figure*}[!htb]
    \centering
    \includegraphics[width=\textwidth]{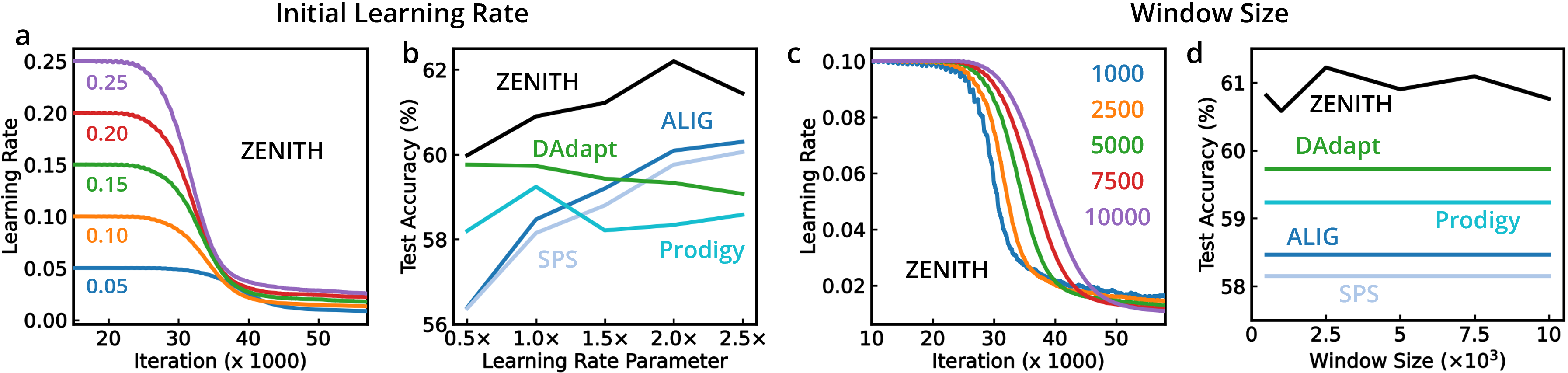} 
    \caption{Hyperparameter Sensitivity Analysis. (a-d) Effect of initial LR and window size parameters on LR trajectories and test accuracies of ZENITH in the Tiny ImageNet setting.}
    \label{fig:hyperparameters}
\end{figure*}

\subsubsection{Minima Sharpness and Generalization.}
ZENITH's high test accuracy is attributed to its ability to maintain a high LR when necessary. Higher LRs evade sharp local minima and converge to flat global minima. To create Figure~\ref{fig:sharpness}(a-b), models were trained using vanilla SGD across many constant LRs from 0.004 to 0.25. The sharpness of the minima was measured using the top eigenvalue ($\lambda_{\max}$) of the Hessian matrix. Figure 3a shows an inverse relationship between LR and sharpness because higher LRs escaped from sharp basins and converged into flatter regions. Figure 3b shows that these higher LRs correspond to higher test accuracy. Therefore, flatter minima were associated with better generalization because the test loss landscape is shifted relative to the training landscape due to distribution mismatches. Consequently, a sharp minimum incurs a larger penalty in test loss than a flat one. This theory aligns with the empirical observations in Figure~\ref{fig:sharpness}c, which depicts the loss landscapes around the minima reached with CIFAR-10 in our main experiments. COCOB, which uses tiny step sizes, gets trapped in a sharper minimum, resulting in a low test accuracy of 89.9\%. Similarly, ALIG and SPS used maximum LRs of 0.00341 and 0.00967, leading to even sharper minima and lower test accuracies of 85.3\% and 87.7\%. Conversely, ZENITH maintains a larger LR when necessary, converging to a flatter minimum with a higher test accuracy of 92.4\%. Other top-performing baselines include PAL (91.0\%) and DoWG (92.4\%), which also converged to flatter minima by using a higher LR of up to 0.25. This inverse correlation between sharpness and test accuracy is also consistent across other schedulers and experiments, as summarized in Figure~\ref{fig:sharpness}d. Figure~\ref{fig:sharpness}c shows that high-performing ZENITH, DoWG, and PAL converged to minima with lower loss, indicating that final training loss influences accuracy too. A detailed formal analysis explaining ZENITH’s better generalization is presented in Appendix~\ref{sec:formal_analysis}B. 

\subsubsection{Hyperparameter Sensitivity Analysis.}
We conducted a grid search over the LR-related parameters. For most schedulers (e.g., Quadratic, Polyak, and ZENITH), this parameter is the initial/maximum LR ($\eta_0$), while for distance-aware estimators, it is the LR growth rate ($\eta_{gr}$). We varied $\eta_0$ from 0.05 to 0.25 in intervals of 0.05, and $\eta_{gr}$ from 0.5 to 2.5 in intervals of 0.5. Both variations span from 0.5$\times$ to 2.5$\times$ the default base values of $\eta_0 = 0.1$ and $\eta_{\text{gr}} = 1.0$. Figure~\ref{fig:hyperparameters}(a-b) shows the parameter sensitivity of different algorithms on Tiny ImageNet. As shown in Figure~\ref{fig:hyperparameters}a, ZENITH effectively decays the LR across all values of $\eta_0$. Figure~\ref{fig:hyperparameters}b shows that the maximum accuracy ZENITH achieves is higher than the maximum accuracies of baselines. The accuracies of ZENITH, ALIG, and SPS increase with $\eta_0$ because decaying the LR from a higher starting value biases training toward flatter minima. The earlier fixed-$\eta_0$ experiments showed that ZENITH outperforms baselines under the standard setup used in this research area, where all methods are evaluated with the same $\eta_0$. These results show that ZENITH still outperforms the baselines even when the LR-related parameters are tuned for all methods. Comprehensive LR-parameter grid searches across datasets are provided in Appendix C. 

ZENITH’s only non-LR parameter is the window size ($W$), which was fixed at 5000 in all main experiments. We did not tune this parameter because parameter-free schedulers are supposed to perform robustly using their default parameter values. To examine sensitivity to $W$, we now repeat the Tiny ImageNet experiment using 5 values of $W$ spanning an entire order of magnitude. Figure~\ref{fig:hyperparameters}(c-d) shows that increasing $W$ slightly delays LR decay, but ZENITH still decays the LR effectively. It outperforms the baselines across the full $W$ spectrum. Baselines appear as flat lines because they use non-LR parameters different from $W$. While Figure~\ref{fig:hyperparameters} focuses only on accuracy, ZENITH also maintains its time advantage over baselines, as discussed previously.

\subsubsection{Compatibility with Regularization.}  
\begin{figure}[!t]
    \centering
    \includegraphics[width=\columnwidth]{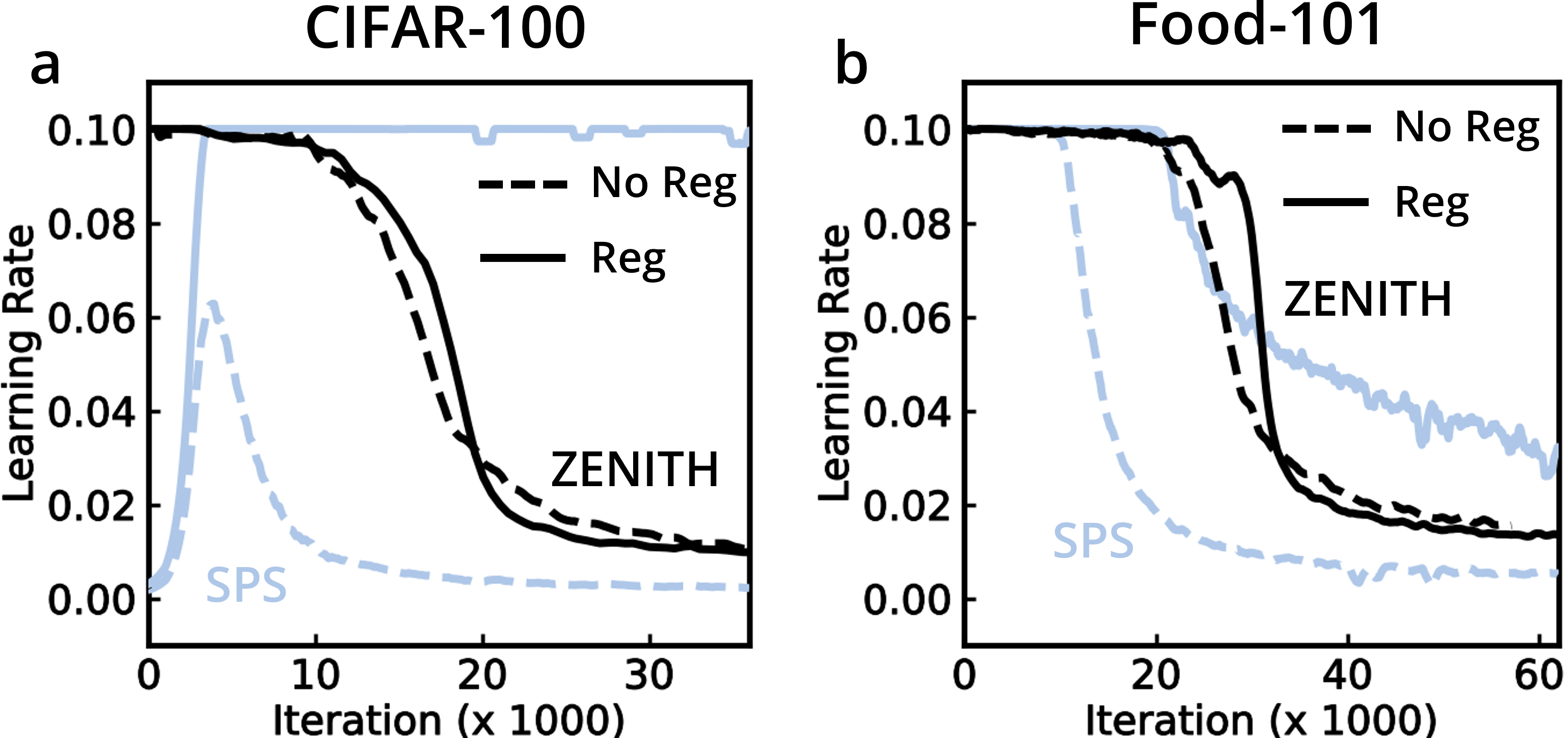}
    \caption{Impact of $L_2$ regularization on LR trajectories.}
    \label{fig:regularization}
\end{figure}

An advantage of ZENITH is its compatibility with regularization. For all experiments so far, we excluded regularization because several baselines were incompatible with it. We now investigate the effect of $L_2$ regularization in the CIFAR-100 and Food-101 settings using the Polyak-style baseline (SPS) and ZENITH. Without regularization, Figure~\ref{fig:regularization}a shows that the scale sensitivity of SPS leads to low LR values below $0.06$ on CIFAR-100. Adding a regularization term of strength $10^{-4}$ increases the training loss numerator of the Polyak step size formula, inflating the LR. However, this causes LR to be constantly overestimated and capped at the maximum value of $0.1$. As a result, training is slower to converge and accuracy remains low. In contrast, ZENITH’s scheduling mechanism remains robust with regularization because it uses reliable gradient signals instead of loss. Hence, regularization improves ZENITH’s generalization. Further regularization experiments involving more baselines are in Appendix D.

\subsubsection{Comparison with Manual Schedules.} 
\begin{figure}[!t]
    \centering
    \includegraphics[width=\columnwidth]{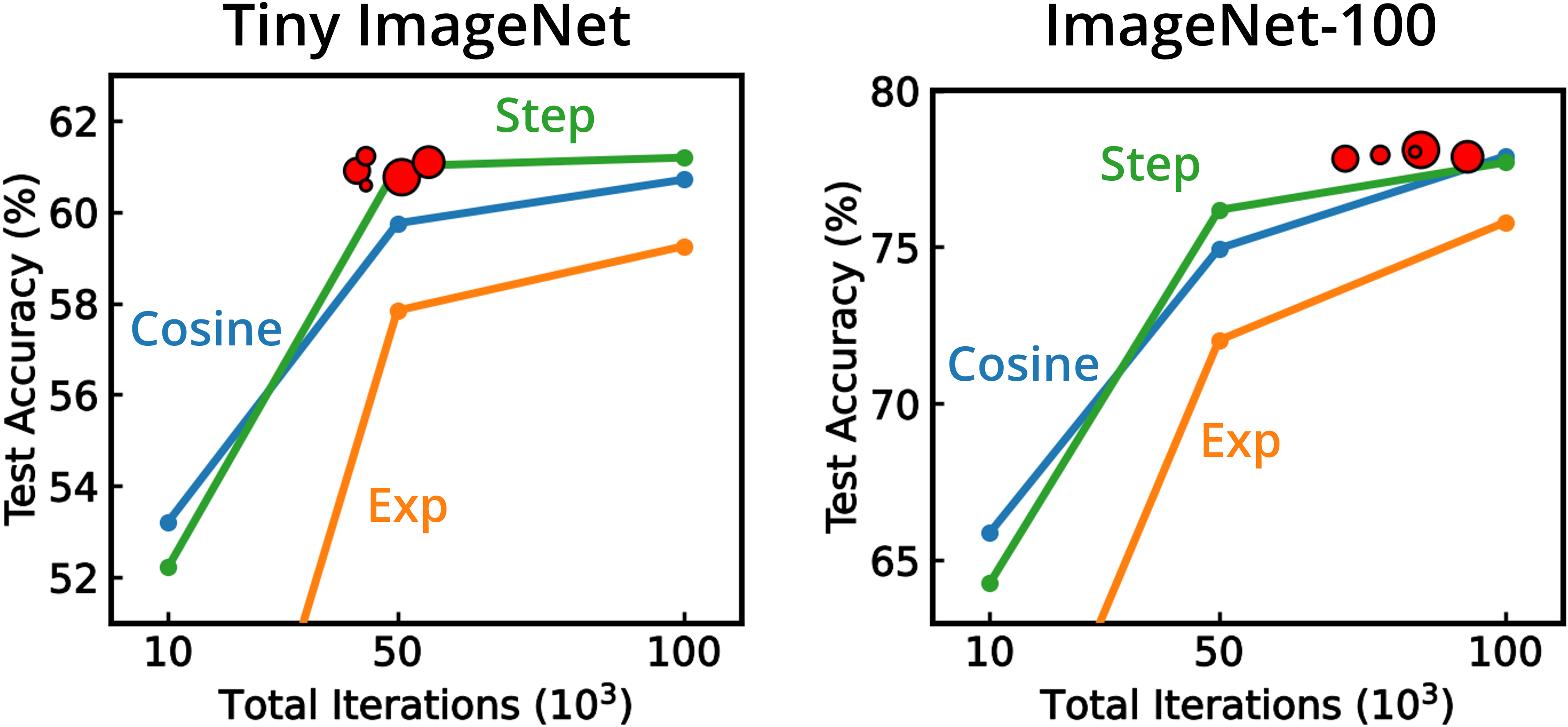}
    \caption{Comparison of parameter sensitivity between manual schedules and ZENITH. The red circles represent ZENITH, with radii proportional to the window size.}
    \label{fig:manual_main_text}
\end{figure}

Cosine annealing, exponential decay, and step decay require the total number of iterations ($N_{\text{iter}}$) as a parameter. We tested three discrete values for $N_{\text{iter}}$: 10,000, 50,000, and 100,000. For cosine annealing, the LR decayed from 0.1 to 0 over $N_{\text{iter}}$ iterations. For exponential decay, the decay factor was set to reduce the LR from 0.1 to $10^{-3}$ by iteration $N_{\text{iter}}$. For step decay, the LR was multiplied by 0.1 at $0.5N_{\text{iter}}$ and $0.75N_{\text{iter}}$. ZENITH instead uses a window size ($W$) as a parameter, which we varied across five values: 1000, 2500, 5000, 7500, and 10,000. The parameter ranges tested for both ZENITH and manual schedules span one order of magnitude. As shown in Figure~\ref{fig:manual_main_text}, manual schedules rely heavily on an optimal choice of $N_{\text{iter}}$, which varies significantly across datasets. For example, ImageNet-100 requires nearly 100,000 iterations to reach high accuracy, whereas running Tiny ImageNet for that long increases training time without yielding benefits. Conversely, underestimating $N_{\text{iter}}$ (i.e. 10,000) results in much lower accuracy. In contrast, ZENITH automatically decays the LR over a suitable timescale regardless of the $W$ chosen. It achieves high test accuracies comparable to tuned manual schedules while avoiding unnecessarily long training periods. The lower accuracy of exponential decay compared to other manual schedules is caused by the premature reduction of the LR in early stages, which supports our analysis of minima flatness. The full comparison with manual schedules across all benchmarks is detailed in Appendix E.

\begin{figure}[!t]
    \centering
    \includegraphics[width=\columnwidth]{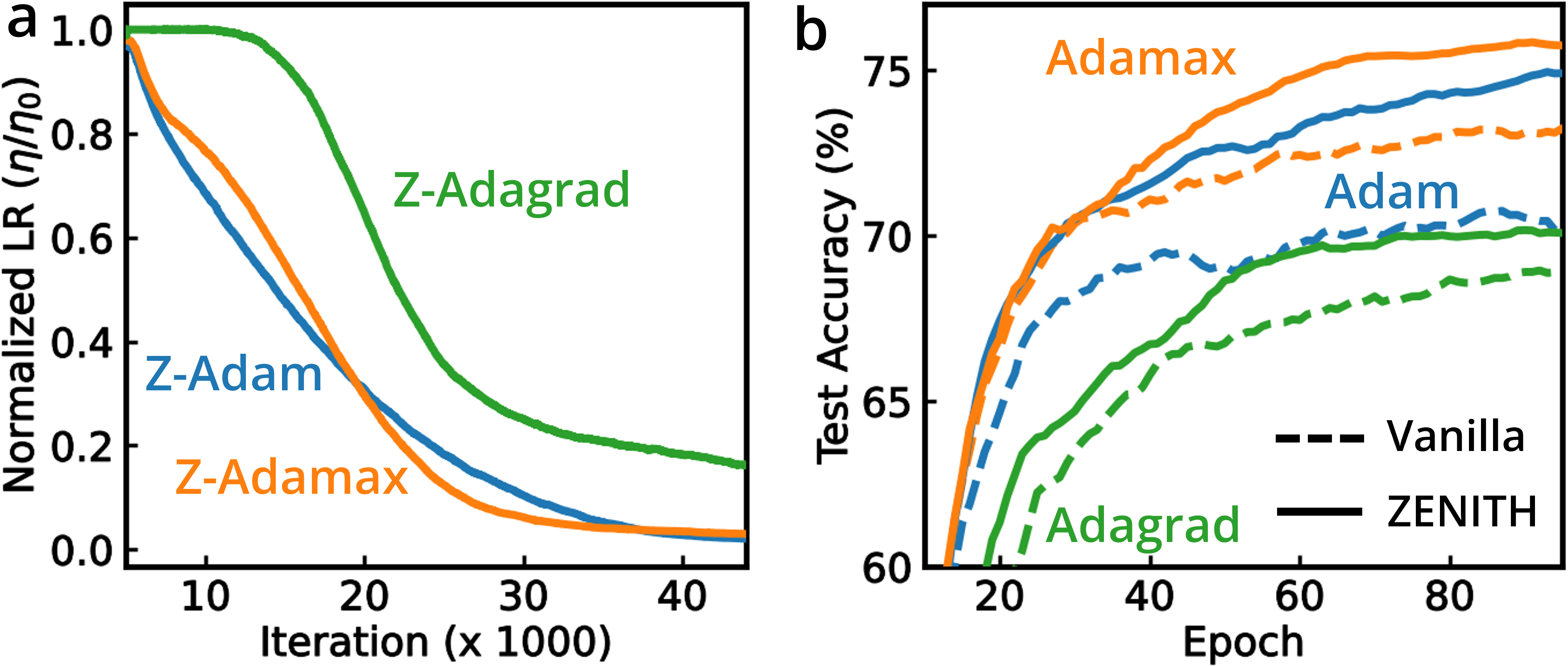}
    \caption{Applying the ZENITH schedule to other base optimizers. (a) LR trajectories and (b) test-accuracy curves.}
    \label{fig:other_optimizer}
\end{figure}

\subsubsection{Other Base Optimizers.}
ZENITH is a scheduler that can be used with other optimizers besides SGD. Figure~\ref{fig:other_optimizer}a shows the LR trajectories generated when applied to Adam, Adamax, and Adagrad on CIFAR-100. Figure~\ref{fig:other_optimizer}b shows that ZENITH enables them to reach higher test accuracy in fewer epochs. While these adaptive optimizers adjust their step sizes dynamically, they rely on short-term signals to maintain immediate stability and often require a separate scheduler for long-term LR decay. Appendix F elaborates on the distinctions between ZENITH and these adaptive optimizers.

\subsubsection{Additional Experiments.} ZENITH's advantages extend to tasks such as detection and segmentation. Unlike image classification where cross-entropy loss can approach zero, detection and segmentation use loss functions that converge to values far above zero. Consequently, Polyak-style methods struggle. Because gradient magnitudes still attenuate toward zero near a minimum, ZENITH's gradient-informed scheduling remains applicable. Experiments on keypoint detection and instance segmentation using Keypoint R-CNN and Mask R-CNN on MS COCO are detailed in Appendix G. We also provide tabular-regression experiments using MLP architectures with Adam as the base optimizer in Appendix H.

\section{Conclusion}
We presented ZENITH, a scheduler that adapts the LR using the history of gradient norms. Comprehensive experiments showed that ZENITH achieves state-of-the-art performance among automatic LR schedulers while reducing wall-clock training time. Further analyses revealed that its success stems from its ability to reach flatter minima while maintaining minimal computational overhead. We validated ZENITH across vision and tabular-data settings, CNN and MLP architectures, and SGD and Adam base optimizers. Future work could explore extensions to other modalities, architectures, and optimizers.

\bibliography{aaai2027}

@inproceedings{szegedy2016rethinking,
  title={Rethinking the Inception Architecture for Computer Vision},
  author={Szegedy, Christian and Vanhoucke, Vincent and Ioffe, Sergey and Shlens, Jon and Wojna, Zbigniew},
  booktitle={Proceedings of the IEEE Conference on Computer Vision and Pattern Recognition},
  pages={2818--2826},
  year={2016}
}

@article{ge2019step,
  title={The Step Decay Schedule: A Near Optimal, Geometrically Decaying Learning Rate Procedure for Least Squares},
  author={Ge, Rong and Kakade, Sham M and Kidambi, Rahul and Netrapalli, Praneeth},
  journal={Advances in Neural Information Processing Systems},
  volume={32},
  year={2019}
}

@misc{loshchilov2016sgdr,
  title={SGDR: Stochastic Gradient Descent with Warm Restarts},
  author={Loshchilov, Ilya and Hutter, Frank},
  year={2016},
  eprint={1608.03983},
  archivePrefix={arXiv},
  primaryClass={cs.LG}
}

@article{orabona2017training,
  title={Training Deep Networks without Learning Rates Through Coin Betting},
  author={Orabona, Francesco and Tommasi, Tatiana},
  journal={Advances in Neural Information Processing Systems},
  volume={30},
  year={2017}
}

@article{mutschler2020parabolic,
  title={Parabolic Approximation Line Search for DNNs},
  author={Mutschler, Maximus and Zell, Andreas},
  journal={Advances in Neural Information Processing Systems},
  volume={33},
  pages={5405--5416},
  year={2020}
}

@article{zhu2021automatic,
  title={Automatic, Dynamic, and Nearly Optimal Learning Rate Specification via Local Quadratic Approximation},
  author={Zhu, Yingqiu and Huang, Danyang and Gao, Yuan and Wu, Rui and Chen, Yu and Zhang, Bo and Wang, Hansheng},
  journal={Neural Networks},
  volume={141},
  pages={11--29},
  year={2021},
  publisher={Elsevier}
}

@inproceedings{fu2024qlabgrad,
  title={QLabGrad: A Hyperparameter-Free and Convergence-Guaranteed Scheme for Deep Learning},
  author={Fu, Minghan and Wu, Fang-Xiang},
  booktitle={Proceedings of the AAAI Conference on Artificial Intelligence},
  volume={38},
  number={11},
  pages={12072--12081},
  year={2024}
}

@misc{bu2024gradient,
  title={Gradient Descent with Generalized Newton's Method},
  author={Bu, Zhiqi and Xu, Shiyun},
  year={2024},
  eprint={2407.02772},
  archivePrefix={arXiv},
  primaryClass={cs.LG}
}

@inproceedings{ivgi2023dog,
  title={DoG is SGD's Best Friend: A Parameter-Free Dynamic Step Size Schedule},
  author={Ivgi, Maor and Hinder, Oliver and Carmon, Yair},
  booktitle={International Conference on Machine Learning},
  pages={14465--14499},
  year={2023},
  organization={PMLR}
}

@article{khaled2023dowg,
  title={DoWG Unleashed: An Efficient Universal Parameter-Free Gradient Descent Method},
  author={Khaled, Ahmed and Mishchenko, Konstantin and Jin, Chi},
  journal={Advances in Neural Information Processing Systems},
  volume={36},
  pages={6748--6769},
  year={2023}
}

@inproceedings{defazio2023learning,
  title={Learning-Rate-Free Learning by D-Adaptation},
  author={Defazio, Aaron and Mishchenko, Konstantin},
  booktitle={International Conference on Machine Learning},
  pages={7449--7479},
  year={2023},
  organization={PMLR}
}

@misc{mishchenko2023prodigy,
  title={Prodigy: An Expeditiously Adaptive Parameter-Free Learner},
  author={Mishchenko, Konstantin and Defazio, Aaron},
  year={2023},
  eprint={2306.06101},
  archivePrefix={arXiv},
  primaryClass={cs.LG}
}

@article{rolinek2018l4,
  title={L4: Practical Loss-Based Stepsize Adaptation for Deep Learning},
  author={Rolinek, Michal and Martius, Georg},
  journal={Advances in Neural Information Processing Systems},
  volume={31},
  year={2018}
}

@inproceedings{berrada2020training,
  title={Training Neural Networks for and by Interpolation},
  author={Berrada, Leonard and Zisserman, Andrew and Kumar, M Pawan},
  booktitle={International Conference on Machine Learning},
  pages={799--809},
  year={2020},
  organization={PMLR}
}

@inproceedings{loizou2021stochastic,
  title={Stochastic Polyak Step-Size for SGD: An Adaptive Learning Rate for Fast Convergence},
  author={Loizou, Nicolas and Vaswani, Sharan and Laradji, Issam Hadj and Lacoste-Julien, Simon},
  booktitle={International Conference on Artificial Intelligence and Statistics},
  pages={1306--1314},
  year={2021},
  organization={PMLR}
}

@article{lecun2002gradient,
  title={Gradient-Based Learning Applied to Document Recognition},
  author={LeCun, Yann and Bottou, L{\'e}on and Bengio, Yoshua and Haffner, Patrick},
  journal={Proceedings of the IEEE},
  volume={86},
  number={11},
  pages={2278--2324},
  year={2002},
  publisher={IEEE}
}

@article{krizhevsky2009learning,
  title={Learning Multiple Layers of Features from Tiny Images},
  author={Krizhevsky, Alex and Hinton, Geoffrey and others},
  journal={Technical Report, University of Toronto},
  year={2009}
}

@inproceedings{bossard2014food,
  title={Food-101 -- Mining Discriminative Components with Random Forests},
  author={Bossard, Lukas and Guillaumin, Matthieu and Van Gool, Luc},
  booktitle={European Conference on Computer Vision},
  pages={446--461},
  year={2014},
  organization={Springer}
}

@article{le2015tiny,
  title={Tiny ImageNet Visual Recognition Challenge},
  author={Le, Yann and Yang, Xuan},
  journal={CS 231N},
  volume={7},
  number={7},
  pages={3},
  year={2015}
}

@inproceedings{deng2009imagenet,
  title={ImageNet: A Large-Scale Hierarchical Image Database},
  author={Deng, Jia and Dong, Wei and Socher, Richard and Li, Li-Jia and Li, Kai and Fei-Fei, Li},
  booktitle={Proceedings of the IEEE Conference on Computer Vision and Pattern Recognition},
  pages={248--255},
  year={2009},
  organization={IEEE}
}

@inproceedings{tan2019efficientnet,
  title={EfficientNet: Rethinking Model Scaling for Convolutional Neural Networks},
  author={Tan, Mingxing and Le, Quoc},
  booktitle={International Conference on Machine Learning},
  pages={6105--6114},
  year={2019},
  organization={PMLR}
}

@misc{simonyan2014very,
  title={Very Deep Convolutional Networks for Large-Scale Image Recognition},
  author={Simonyan, Karen and Zisserman, Andrew},
  year={2014},
  eprint={1409.1556},
  archivePrefix={arXiv},
  primaryClass={cs.CV}
}

@inproceedings{he2016deep,
  title={Deep Residual Learning for Image Recognition},
  author={He, Kaiming and Zhang, Xiangyu and Ren, Shaoqing and Sun, Jian},
  booktitle={Proceedings of the IEEE Conference on Computer Vision and Pattern Recognition},
  pages={770--778},
  year={2016}
}

@inproceedings{yu2018deep,
  title={Deep Layer Aggregation},
  author={Yu, Fisher and Wang, Dequan and Shelhamer, Evan and Darrell, Trevor},
  booktitle={Proceedings of the IEEE Conference on Computer Vision and Pattern Recognition},
  pages={2403--2412},
  year={2018}
}

@inproceedings{huang2017densely,
  title={Densely Connected Convolutional Networks},
  author={Huang, Gao and Liu, Zhuang and Van Der Maaten, Laurens and Weinberger, Kilian Q},
  booktitle={Proceedings of the IEEE Conference on Computer Vision and Pattern Recognition},
  pages={4700--4708},
  year={2017}
}

@inproceedings{lin2014microsoft,
  title={Microsoft COCO: Common Objects in Context},
  author={Lin, Tsung-Yi and Maire, Michael and Belongie, Serge and Hays, James and Perona, Pietro and Ramanan, Deva and Doll{\'a}r, Piotr and Zitnick, C Lawrence},
  booktitle={European Conference on Computer Vision},
  pages={740--755},
  year={2014},
  organization={Springer}
}

@inproceedings{he2017mask,
  title={Mask R-CNN},
  author={He, Kaiming and Gkioxari, Georgia and Doll{\'a}r, Piotr and Girshick, Ross},
  booktitle={Proceedings of the IEEE International Conference on Computer Vision},
  pages={2961--2969},
  year={2017}
}

@inproceedings{keskar2017large,
  title={On Large-Batch Training for Deep Learning: Generalization Gap and Sharp Minima},
  author={Keskar, Nitish Shirish and Mudigere, Dheevatsa and Nocedal, Jorge and Smelyanskiy, Mikhail and Tang, Ping Tak Peter},
  booktitle={International Conference on Learning Representations},
  year={2017}
}

@misc{dua2019individual,
  title={Individual Household Electric Power Consumption Data Set},
  author={Dua, Dheeru and Graff, Casey},
  year={2019},
  howpublished={UCI Machine Learning Repository},
  url={https://archive.ics.uci.edu/ml/datasets/individual+household+electric+power+consumption}
}

@inproceedings{bertin2011million,
  title={The Million Song Dataset},
  author={Bertin-Mahieux, Thierry and Ellis, Daniel P.W. and Whitman, Brian and Lamere, Paul},
  booktitle={Proceedings of the 12th International Society for Music Information Retrieval Conference (ISMIR)},
  pages={591--596},
  year={2011}
}

\appendix

\twocolumn[{%
  \begin{center}
    \LARGE \textbf{Appendix}
  \end{center}

When interpreting the tables, it is important to note that they provide only a limited view of comparative performance. The figures offer a more complete comparison because they show accuracy at every wall-clock time point throughout training. In contrast, the tables report only the best observed accuracy--time trade-off, which cannot fully capture how each algorithm performs over the entire training process.

  \section{A. Tables of Experimental Results}
  \label{sec:appendix_tables} 


\begin{center}

    \vspace{0.75em}
    \label{tab:optimizer_comparison_mean_std_part1}
    
    \setlength{\aboverulesep}{0pt}
    \setlength{\belowrulesep}{0pt}
    \renewcommand{\arraystretch}{1.15}
    
    \begin{NiceTabularX}{\textwidth}{l | Y Y Y Y Y Y}
        \CodeBefore
            \rowcolors{3}{lightgray}{white}
        \Body
        \hline 
        \Block{2-1}{Method} & 
        \multicolumn{2}{c}{MNIST} &
        \multicolumn{2}{c}{CIFAR-10} & 
        \multicolumn{2}{c}{CIFAR-100} \\
        \cline{2-7} 
         & \makecell{Acc (\%)} & \makecell{Time (min)} 
         & \makecell{Acc (\%)} & \makecell{Time (min)} 
         & \makecell{Acc (\%)} & \makecell{Time (min)} \\
        \hline 
        SGD & $99.62 \pm 0.01$ & $8.54 \pm 3.40$ & $92.69 \pm 0.26$ & $28.42 \pm 11.65$ & $74.32 \pm 0.20$ & $49.40 \pm 4.15$ \\    
        \hline
        PAL & $99.60 \pm 0.08$ & $17.13 \pm 4.88$ & $90.96 \pm 0.24$ & $15.00 \pm 5.19$ & $68.68 \pm 0.35$ & $29.21 \pm 7.83$ \\
        LQA & $99.59 \pm 0.02$ & $57.13 \pm 2.32$ & $89.03 \pm 0.07$ & $32.93 \pm 1.01$ & $63.40 \pm 1.81$ & $58.89 \pm 0.79$ \\
        QLabGrad & $99.59 \pm 0.05$ & $9.21 \pm 5.08$ & $92.29 \pm 0.34$ & $18.30 \pm 3.65$ & $72.67 \pm 0.23$ & $44.92 \pm 9.00$ \\
        GeN & $99.32 \pm 0.11$ & $4.54 \pm 2.23$ & $78.70 \pm 0.59$ & $33.59 \pm 2.27$ & $45.22 \pm 1.36$ & $59.11 \pm 0.91$ \\
        \hline
        COCOB & $99.60 \pm 0.03$ & $9.99 \pm 1.19$ & $89.89 \pm 0.05$ & $17.68 \pm 2.90$ & $65.37 \pm 0.37$ & $55.90 \pm 4.11$ \\
        \hline
        DoG & $99.56 \pm 0.04$ & $15.81 \pm 2.53$ & $84.99 \pm 0.53$ & $31.86 \pm 2.60$ & $14.23 \pm 11.67$ & $39.94 \pm 34.22$ \\
        DoWG & $99.56 \pm 0.05$ & $10.62 \pm 0.33$ & $92.59 \pm 0.06$ & $26.41 \pm 4.88$ & $71.65 \pm 1.56$ & $41.31 \pm 13.24$ \\
        DAdapt & $99.58 \pm 0.06$ & $7.08 \pm 1.48$ & $89.75 \pm 0.13$ & $29.45 \pm 5.11$ & $68.90 \pm 0.52$ & $54.51 \pm 5.50$ \\
        Prodigy & $99.57 \pm 0.07$ & $17.93 \pm 10.34$ & $89.47 \pm 0.16$ & $25.33 \pm 1.25$ & $67.74 \pm 0.45$ & $49.21 \pm 4.65$ \\
        \hline
        L4 & $99.42 \pm 0.07$ & $10.52 \pm 4.03$ & $83.44 \pm 0.29$ & $25.53 \pm 3.59$ & $58.00 \pm 0.15$ & $57.80 \pm 1.64$ \\
        ALIG & $99.50 \pm 0.02$ & $10.77 \pm 2.45$ & $85.18 \pm 0.38$ & $19.29 \pm 16.99$ & $67.32 \pm 0.32$ & $22.42 \pm 16.60$ \\
        SPS & $99.48 \pm 0.07$ & $11.05 \pm 2.96$ & $87.82 \pm 0.10$ & $13.39 \pm 1.61$ & $69.75 \pm 0.34$ & $34.66 \pm 6.64$ \\
        \hline
        \textbf{ZENITH} & $\underline{\boldsymbol{99.64 \pm 0.01}}$ & $\underline{\boldsymbol{5.57 \pm 1.80}}$ & $\underline{\boldsymbol{92.91 \pm 0.17}}$ & $\underline{\boldsymbol{15.84 \pm 2.85}}$ & $\underline{\boldsymbol{74.54 \pm 0.15}}$ & $\underline{\boldsymbol{33.40 \pm 1.39}}$ \\
        \hline 
    \end{NiceTabularX}

    \captionof{table}{Comparison of test accuracy and wall-clock time for MNIST, CIFAR-10, and CIFAR-100. Results are reported as Mean $\pm$ Standard Deviation over 3 runs. The best combinations of accuracy and time results are bold and underlined. A dash indicates that the method failed to learn. This table shows the best instantaneous combinations of accuracy and requisite training time to achieve the accuracy. For a complete view of how performance evolves over time, please refer to the learning curves in Figure~\ref{fig:image_classification} of the main text. The figures provide a more holistic and accurate view of the training efficacy of the competing algorithms.}

    \vspace{1em} 


    \vspace{0.75em}
    \label{tab:optimizer_comparison_mean_std_part2}
    
    \setlength{\aboverulesep}{0pt}
    \setlength{\belowrulesep}{0pt}
    \renewcommand{\arraystretch}{1.15}
    
    \begin{NiceTabularX}{\textwidth}{l | Y Y Y Y Y Y}
        \CodeBefore
            \rowcolors{3}{lightgray}{white}
        \Body
        \hline 
        \Block{2-1}{Method} & 
        \multicolumn{2}{c}{Food-101} &
        \multicolumn{2}{c}{Tiny ImageNet} & 
        \multicolumn{2}{c}{ImageNet-100} \\
        \cline{2-7} 
         & Acc (\%) & Time (hr) 
         & Acc (\%) & Time (hr) 
         & Acc (\%) & Time (hr) \\
        \hline 
        SGD & $74.68 \pm 0.16$ & $4.52 \pm 0.37$ & $60.28 \pm 0.12$ & $4.95 \pm 0.19$ & $77.87 \pm 0.46$ & $8.21 \pm 1.47$ \\         
        \hline 
        PAL & $68.58 \pm 0.24$ & $0.86 \pm 0.19$ & $57.83 \pm 0.13$ & $2.52 \pm 0.21$ & $72.02 \pm 0.59$ & $3.67 \pm 2.15$ \\
        LQA & $73.39 \pm 0.71$ & $2.38 \pm 0.30$ & $58.62 \pm 0.54$ & $4.14 \pm 0.58$ & — & — \\
        QLabGrad & — & — & — & — & $75.72 \pm 0.58$ & $3.81 \pm 3.57$ \\
        GeN & $44.88 \pm 0.01$ & $4.98 \pm 0.02$ & $31.47 \pm 0.97$ & $4.72 \pm 0.20$ & $54.93 \pm 0.51$ & $9.74 \pm 0.09$ \\
        \hline
        COCOB & $63.97 \pm 0.35$ & $1.21 \pm 0.33$ & $40.71 \pm 2.13$ & $2.46 \pm 0.92$ & $68.55 \pm 0.31$ & $3.38 \pm 0.88$ \\
        \hline
        DoG & $68.17 \pm 0.36$ & $3.66 \pm 0.81$ & $52.14 \pm 6.99$ & $3.72 \pm 1.62$ & $76.52 \pm 0.54$ & $7.64 \pm 0.77$ \\
        DoWG & $68.98 \pm 0.80$ & $1.54 \pm 0.20$ & $51.13 \pm 4.60$ & $3.12 \pm 1.96$ & $76.14 \pm 0.40$ & $7.98 \pm 0.70$ \\
        DAdapt & $73.29 \pm 0.66$ & $2.75 \pm 0.48$ & $59.73 \pm 0.54$ & $4.81 \pm 0.14$ & $77.21 \pm 0.20$ & $8.25 \pm 0.62$ \\
        Prodigy & $71.56 \pm 0.83$ & $1.85 \pm 0.53$ & $59.24 \pm 0.61$ & $4.19 \pm 0.85$ & $77.99 \pm 0.45$ & $6.57 \pm 2.50$ \\
        \hline
        L4 & — & — & — & — & $69.97 \pm 2.29$ & $3.56 \pm 1.13$ \\
        ALIG & $72.27 \pm 0.09$ & $0.71 \pm 0.03$ & $58.47 \pm 0.49$ & $1.53 \pm 0.19$ & $74.76 \pm 0.26$ & $1.71 \pm 0.15$ \\
        SPS & $73.72 \pm 0.16$ & $1.07 \pm 0.32$ & $58.15 \pm 0.65$ & $2.15 \pm 0.33$ & $76.56 \pm 0.47$ & $1.78 \pm 0.11$ \\
        \hline
        \textbf{ZENITH} & $\underline{\boldsymbol{75.15 \pm 0.03}}$ & $\underline{\boldsymbol{1.58 \pm 0.24}}$ & $\underline{\boldsymbol{60.90 \pm 0.22}}$ & $\underline{\boldsymbol{2.87 \pm 0.05}}$ & $\underline{\boldsymbol{78.09 \pm 0.23}}$ & $\underline{\boldsymbol{4.36 \pm 0.62}}$ \\
        \hline 
    \end{NiceTabularX}
    \captionof{table}{Comparison of test accuracy and requisite wall-clock time for Food-101, Tiny ImageNet, and ImageNet-100. Results are reported as Mean $\pm$ Standard Deviation over 3 runs. The best combinations of accuracy and time results are bold and underlined. A dash indicates that the method failed to learn. This table shows the best instantaneous combinations of accuracy and requisite training time to achieve the accuracy. For a complete view of how performance evolves over time, please refer to the learning curves in Figure~\ref{fig:image_classification} of the main text. The figures provide a more holistic and accurate view of the training efficacy of the competing algorithms.}

\end{center}
}] 

\twocolumn[{%
    \begin{center}

        \label{tab:optimizer_comparison_average}
        
        \setlength{\aboverulesep}{0pt}
        \setlength{\belowrulesep}{0pt}
        \renewcommand{\arraystretch}{1.15}
        
        \begin{NiceTabular}{l | c c}
            \CodeBefore
                \rowcolors{3}{lightgray}{white}
            \Body
            \hline 
            \textbf{Method} & \textbf{Normalized Acc (vs SGD)} & \textbf{Normalized Time (vs SGD)} \\
            \hline 
            L4 & $0.52\text{x}$ & $0.83\text{x}$ \\
            QLabGrad & $0.59\text{x}$ & $0.67\text{x}$ \\
            GeN & $0.66\text{x}$ & $1.12\text{x}$ \\
            LQA & $0.75\text{x}$ & $0.93\text{x}$ \\
            \hline
            DoG & $0.77\text{x}$ & $0.89\text{x}$ \\
            COCOB & $0.85\text{x}$ & $0.59\text{x}$ \\
            PAL & $0.94\text{x}$ & $0.45\text{x}$ \\
            DoWG & $0.94\text{x}$ & $0.74\text{x}$ \\
            \hline
            ALIG & $0.94\text{x}$ & $0.36\text{x}$ \\
            Prodigy & $0.96\text{x}$ & $0.79\text{x}$ \\
            SPS & $0.96\text{x}$ & $0.41\text{x}$ \\
            DAdapt & $0.97\text{x}$ & $0.95\text{x}$ \\
            \hline
            Vanilla SGD & $1.00\text{x}$ & $1.00\text{x}$ \\         
            \textbf{ZENITH} & $\underline{\boldsymbol{1.01\text{x}}}$ & $\underline{\boldsymbol{0.54\text{x}}}$ \\
            \hline 
        \end{NiceTabular}
        \captionof{table}{Summary of normalized test accuracy and wall-clock time across the five non-MNIST image-classification datasets (CIFAR-10, CIFAR-100, Food-101, Tiny ImageNet, ImageNet-100). MNIST was excluded because it is too simple; even vanilla SGD reaches peak accuracy, meaning it does not reflect the complexity of real-world tasks. Algorithms that failed to learn on a dataset were assigned an accuracy of 0.00x for that dataset, while their time averages were computed only over their completed runs. The best combination of accuracy and time is bold and underlined.}

    \end{center}

\vspace{1.5em} 
}] 


\section{B. Formal Analysis: ZENITH's Better Generalization}
\label{sec:formal_analysis} 

This section provides a formal theoretical analysis to support the conceptual and empirical analysis from the main text on how ZENITH converges to flatter minima and generalizes better. 

\subsection{Why does ZENITH reach flatter minima?}
Let $H = \nabla^2\mathcal{L}(\theta^*)$ be the Hessian matrix evaluated at a minimum point $\theta^*$. The sharpness of this point is quantified by its dominant eigenvalue $\lambda_{\max}$. We can approximate the gradient near $\theta^*$ using a first-order Taylor expansion:
\begin{equation}\label{eq:grad_approx}
    \nabla\mathcal{L}(\theta_t) \approx H(\theta_t - \theta^*)
\end{equation}
The gradient descent update rule is:
\begin{equation}\label{eq:gd_update}
    \theta_{t+1} = \theta_t - \eta_t \nabla\mathcal{L}(\theta_t)
\end{equation}
Substituting equation \eqref{eq:grad_approx} into equation \eqref{eq:gd_update} yields:
\begin{equation}\label{eq:subbed_update}
    \theta_{t+1} - \theta^* = \theta_t - \theta^* - \eta_t H(\theta_t - \theta^*)
\end{equation}

Let $e_t = \theta_t - \theta^*$ within equation \eqref{eq:subbed_update} represent the error vector because it represents the distance of $\theta_t$ from the minimum point $\theta^*$. Substituting $e_t$ and $e_{t+1}$ into equation \eqref{eq:subbed_update} yields the following discrete-time linear dynamical system:
\begin{equation}\label{eq:error_decay}
    e_{t+1} = (I - \eta_t H)e_t
\end{equation}

For the local quadratic dynamics to contract toward this minimum, the error $e_t$ must decay across iterations. This requires the spectral radius of the transition matrix $(I - \eta_t H)$ within equation \eqref{eq:error_decay} to be strictly bounded by 1:
\begin{equation}\label{eq:stability_cond}
    |1 - \eta_t \lambda_i| < 1 \quad \forall i
\end{equation}
\begin{equation}
    -1 < 1 - \eta_t \lambda_i < 1
\end{equation}
\begin{equation}\label{eq:eta_bound}
    \eta_t < \frac{2}{\lambda_{\max}} \implies \lambda_{\max} < \frac{2}{\eta_t}
\end{equation}
Therefore, equation \eqref{eq:eta_bound} states that $\eta_t$ must satisfy this condition across all eigenvalues $\lambda_i$ to remain within a minimum. Hence, gradient descent has a propensity to escape minima whose sharpness $\lambda_{\max}$ exceeds $\frac{2}{\eta_t}$. ZENITH maintains a higher $\eta_t$ during the exploratory training phase, which means that the stability threshold $\frac{2}{\eta_t}$ is low. Consequently, the condition \eqref{eq:eta_bound} is not satisfied in sharper regions, and thus ZENITH tends to traverse the loss landscape until it finds a basin flat enough to satisfy the condition. This promotes convergence to flatter minima than baselines like COCOB and Polyak-style schedulers, which often prematurely drop $\eta_t$, satisfying the condition \eqref{eq:eta_bound} even inside sharp minima.

\subsection{Why do flatter minima improve test accuracy?}
Let $\theta^*$ be the optimal weights of the training loss $\mathcal{L}_{train}$, at which $\nabla \mathcal{L}_{train} = 0$. Since the training and test data are drawn from the same underlying distribution, the test loss $\mathcal{L}_{test}$ shares approximately the same local curvature (Hessian $H$) as the training loss $\mathcal{L}_{train}$ \cite{keskar2017large}. However, finite-sample effects can make their empirical loss landscapes differ slightly. We model this mismatch as a perturbation $\Delta\theta$ in the location of the corresponding optimum:
\begin{equation}\label{eq:dist_mis}
    \mathcal{L}_{test}(\theta^*) \approx \mathcal{L}_{train}(\theta^* + \Delta\theta)
\end{equation}
We apply a Taylor expansion to $\mathcal{L}_{train}$ around the minimum $\theta^*$, evaluated at the shifted parameters $\theta^*+ \Delta\theta$:
\begin{equation}\label{eq:loss_taylor}
\begin{split}
    \mathcal{L}_{train}(\theta^* + \Delta\theta) - \mathcal{L}_{train}(\theta^*) \approx \\ 
    \nabla \mathcal{L}_{train}(\theta^*)^T \Delta\theta + \frac{1}{2} \Delta\theta^T H \Delta\theta
\end{split}
\end{equation}

Since $\nabla \mathcal{L}_{train}(\theta^*) = 0$, the first-order term vanishes in equation \eqref{eq:loss_taylor}. Hence, the generalization gap is governed by the quadratic term:
\begin{equation}\label{eq:loss_quadratic}
    \mathcal{L}_{train}(\theta^* + \Delta\theta) - \mathcal{L}_{train}(\theta^*) \approx \frac{1}{2} \Delta\theta^T H \Delta\theta
\end{equation}

Using the Rayleigh quotient, we upper-bound the right-hand side of equation \eqref{eq:loss_quadratic} by the sharpness of the minimum point:
\begin{equation}\label{eq:rayleigh_bound}
    \mathcal{L}_{train}(\theta^* + \Delta\theta) - \mathcal{L}_{train}(\theta^*) \le \frac{1}{2} \lambda_{\max} \|\Delta\theta\|^2
\end{equation}

By invoking equation \eqref{eq:dist_mis}, we can substitute the shifted training loss on the left-hand side with the test loss evaluated at the minimum. This yields our final generalization bound:
\begin{equation}\label{eq:generalization_bound}
    \mathcal{L}_{test}(\theta^*) - \mathcal{L}_{train}(\theta^*) \le \frac{1}{2} \lambda_{\max} \|\Delta\theta\|^2
\end{equation}

The RHS of Equation \eqref{eq:generalization_bound} represents the loss penalty of the distribution shift. Therefore, flatter minima (smaller $\lambda_{\max}$) reduce the penalty of $\Delta\theta$, resulting in lower test loss and higher test accuracy. Conversely, sharper minima (higher $\lambda_{\max}$) increase the penalty of $\Delta\theta$, resulting in higher test loss and lower test accuracy.

\twocolumn[{%

\section{C. Hyperparameter Sensitivity Analysis}
\label{sec:sensitivity} 

\vspace{1.5em} 
Appendix C extends the hyperparameter sensitivity analysis from the main text to additional datasets. Figure~\ref{fig:lr_param_accuracy} compares tuned test accuracy across scheduler-specific hyperparameters, while Figure~\ref{fig:lr_param_trajectory} shows that ZENITH effectively decays the LR across different initial LR values. 

\vspace{1.5em}

\begin{center}
    \includegraphics[width=\textwidth]{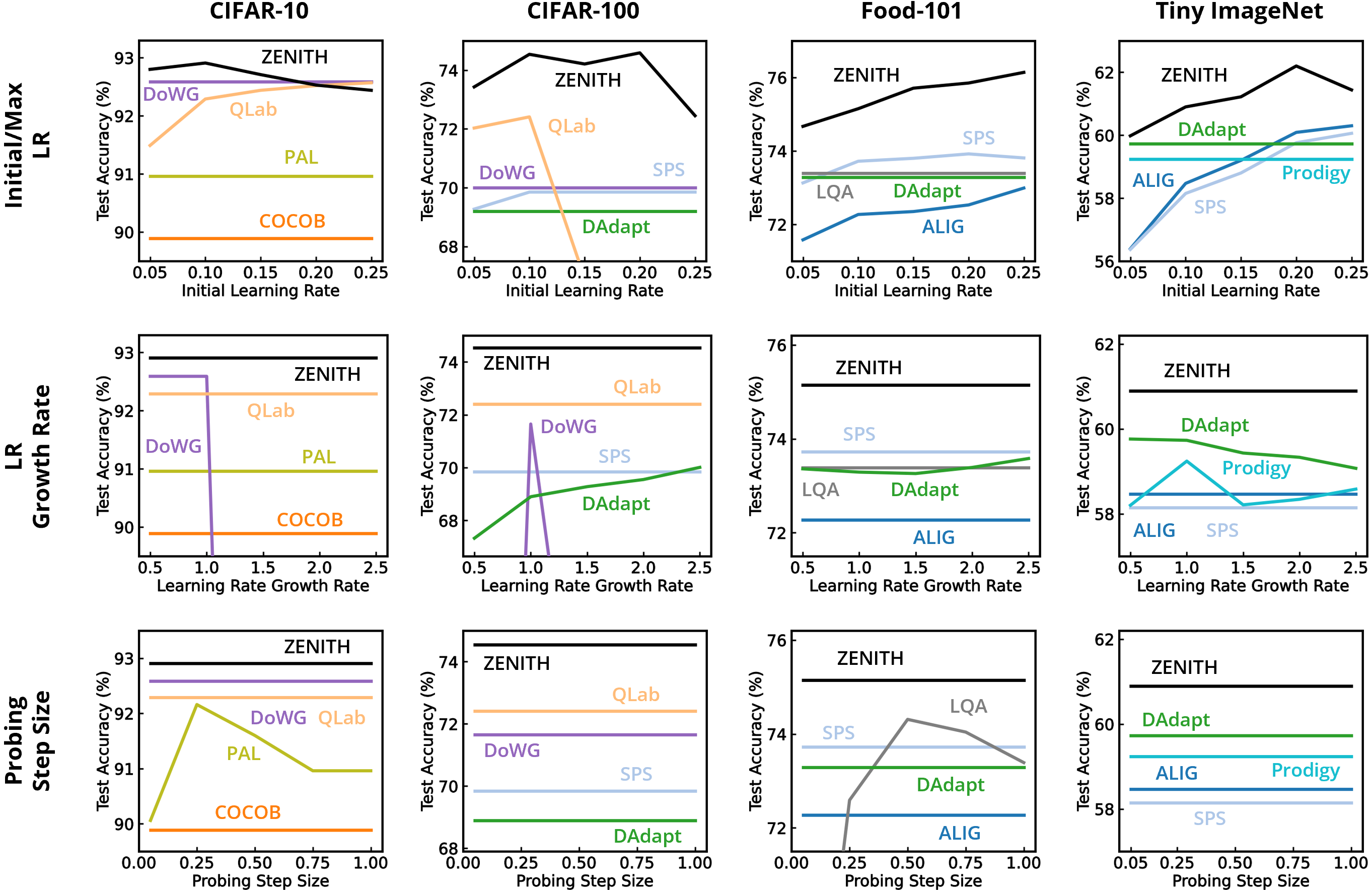}
    \captionof{figure}{Impact of the learning rate parameter on test accuracy. LR-parameter grid searches are conducted across four benchmarks. For most schedulers, the tuned parameter is the initial/maximum LR $\eta_0$; for distance-aware estimators, it is the LR growth rate $\eta_{\mathrm{gr}}$; and for quadratic approximation methods, there is also the additional probing step size. The tuned performance of ZENITH is better than the tuned performance of baselines. Methods without the parameter varied in a given row are shown as horizontal reference lines at their default settings.}
    \label{fig:lr_param_accuracy}
    
    \vspace{2em} 
    
    \includegraphics[width=\textwidth]{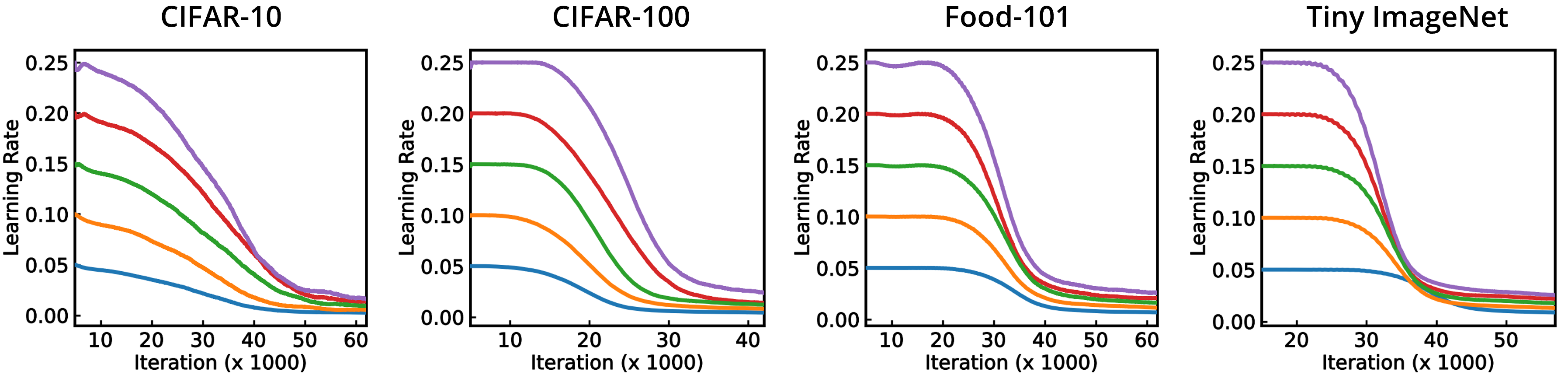}
    \captionof{figure}{LR trajectories generated by ZENITH across four benchmarks for different initial learning rates $\eta_0 \in \{0.05, 0.10, 0.15, 0.20, 0.25\}$.}
    \label{fig:lr_param_trajectory}
\end{center}

\vspace{1.5em} 

}] 


\twocolumn[{%
    \begin{center}
    
    \section{D. Experiments with Regularization}
    \label{sec:regularization} 

    \vspace{1em} 


    \label{tab:optimizer_comparison_reg}
    
    \setlength{\aboverulesep}{0pt}
    \setlength{\belowrulesep}{0pt}
    \renewcommand{\arraystretch}{1.15}

    \vspace{1em}
    
    \begin{NiceTabularX}{\textwidth}{l | Y Y Y Y Y Y | Y Y Y Y Y Y}
        \CodeBefore
            \rowcolors{4}{white}{lightgray}
        \Body
        
        \hline 
        
        \Block{3-1}{Method} & 
        \multicolumn{6}{c|}{\makecell{CIFAR-100}} & 
        \multicolumn{6}{c}{\makecell{Food-101}} \\
        
        \cline{2-7} \cline{8-13} 
        
         & \multicolumn{2}{c}{$1\times10^{-4}$} & \multicolumn{2}{c}{$2\times10^{-4}$} & \multicolumn{2}{c}{$3\times10^{-4}$}
         & \multicolumn{2}{c}{$1\times10^{-4}$} & \multicolumn{2}{c}{$2\times10^{-4}$} & \multicolumn{2}{c}{$3\times10^{-4}$} \\
        
        \cline{2-3} \cline{4-5} \cline{6-7} 
        \cline{8-9} \cline{10-11} \cline{12-13}
        
         & \makecell{Acc\\(\%)} & \makecell{Time\\(min)} 
         & \makecell{Acc\\(\%)} & \makecell{Time\\(min)} 
         & \makecell{Acc\\(\%)} & \makecell{Time\\(min)} 
         & \makecell{Acc\\(\%)} & \makecell{Time\\(hr)}
         & \makecell{Acc\\(\%)} & \makecell{Time\\(hr)}
         & \makecell{Acc\\(\%)} & \makecell{Time\\(hr)} \\
        
        \hline 
        
        DoWG   & 68.35 & 58.0 & 66.85 & 56.1 & 63.47 & 54.2 & 70.75 & 1.57 & 67.84 & 1.42 & 48.10 & 1.92\\
        DAdapt & 62.90 & 56.5 & 59.95 & 44.4 & 50.21 & 58.4 & 69.84 & 1.42 & 52.62 & 0.59 & 18.26 & 0.61\\
        \hline 
        
        ALIG & 74.13 & 59.7 & 74.69 & 55.4 & 74.76 & 46.3 & 72.80 & 1.23 & 73.16 & 1.00 & 74.10 & 0.91 \\
        SPS  & 73.91 & 55.8 & 75.18 & 51.7 & 76.95 & 59.9 & 75.86 & 3.40 & 77.21 & 3.27 & 76.80 & 1.75 \\
        \hline
        \textbf{ZENITH} & \B{74.90} & \B{34.9} & \B{76.06} & \B{30.6} & \B{76.96} & \B{35.1} & \B{76.42} & \B{1.61} & \B{77.70} & \B{2.15} & \B{78.26} & \B{1.65} \\
        
        \hline 
    \end{NiceTabularX}
    \captionof{table}{Comparison of test accuracy and requisite wall-clock time with different regularization magnitudes. This table compares ZENITH to the best baselines on CIFAR-100 and Food-101. Best combinations of accuracy and the requisite time to achieve the accuracy are in bold. It demonstrates that including regularization enhances ZENITH's test accuracy relative to its non-regularized accuracies in Appendix A. Furthermore, increasing the regularization strength from $1\times10^{-4}$ to $3\times10^{-4}$ progressively improved ZENITH's test performance. Distance-aware optimizers showed degraded performance when regularization was introduced compared to the non-regularized setting. This is because regularization can cause their LR estimates to grow too rapidly to maintain stability.}

    \vspace{1.5em} 

    \section{E. Extended Comparison with Manual Schedules}
    \label{sec:appendix_manual_extended} 

    \vspace{1em} 
    
    \includegraphics[width=0.9\textwidth]{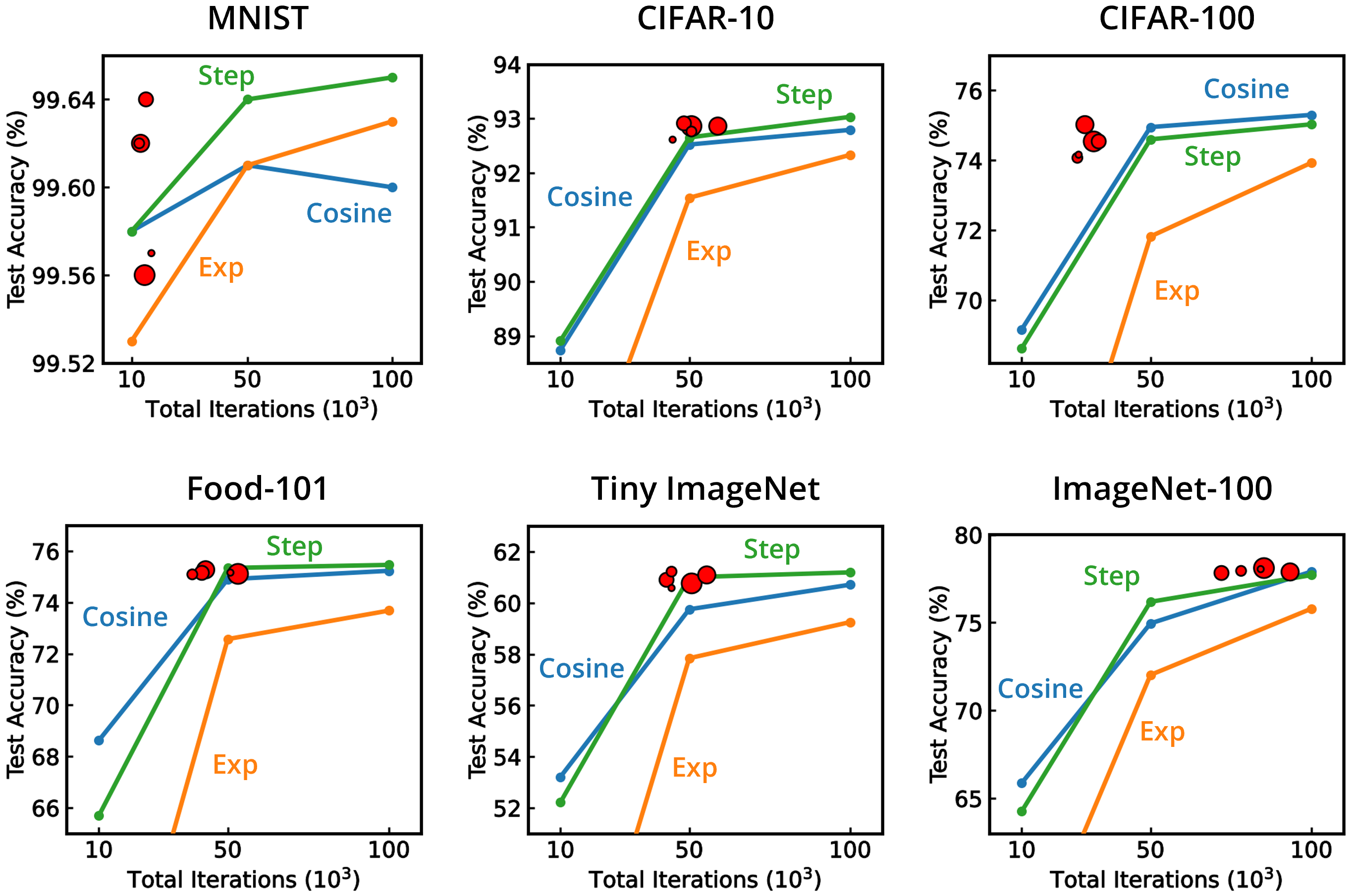}
    \captionof{figure}{Extended comparison of parameter sensitivity for manual schedules and ZENITH across all six evaluated datasets. Manual schedules are varied over $N_{\text{iter}}$, while ZENITH is varied over the window size $W$. Red circles represent ZENITH, with radii proportional to $W$.}
    \label{fig:manual_appendix}

    \end{center}
}] 

\twocolumn[{%
    \begin{center}
    
    \section{F. Distinction between ZENITH and Adaptive Optimizers}
    \label{sec:appendix_distinctions} 

    \vspace{1em} 
    
    \includegraphics[width=0.75\textwidth]{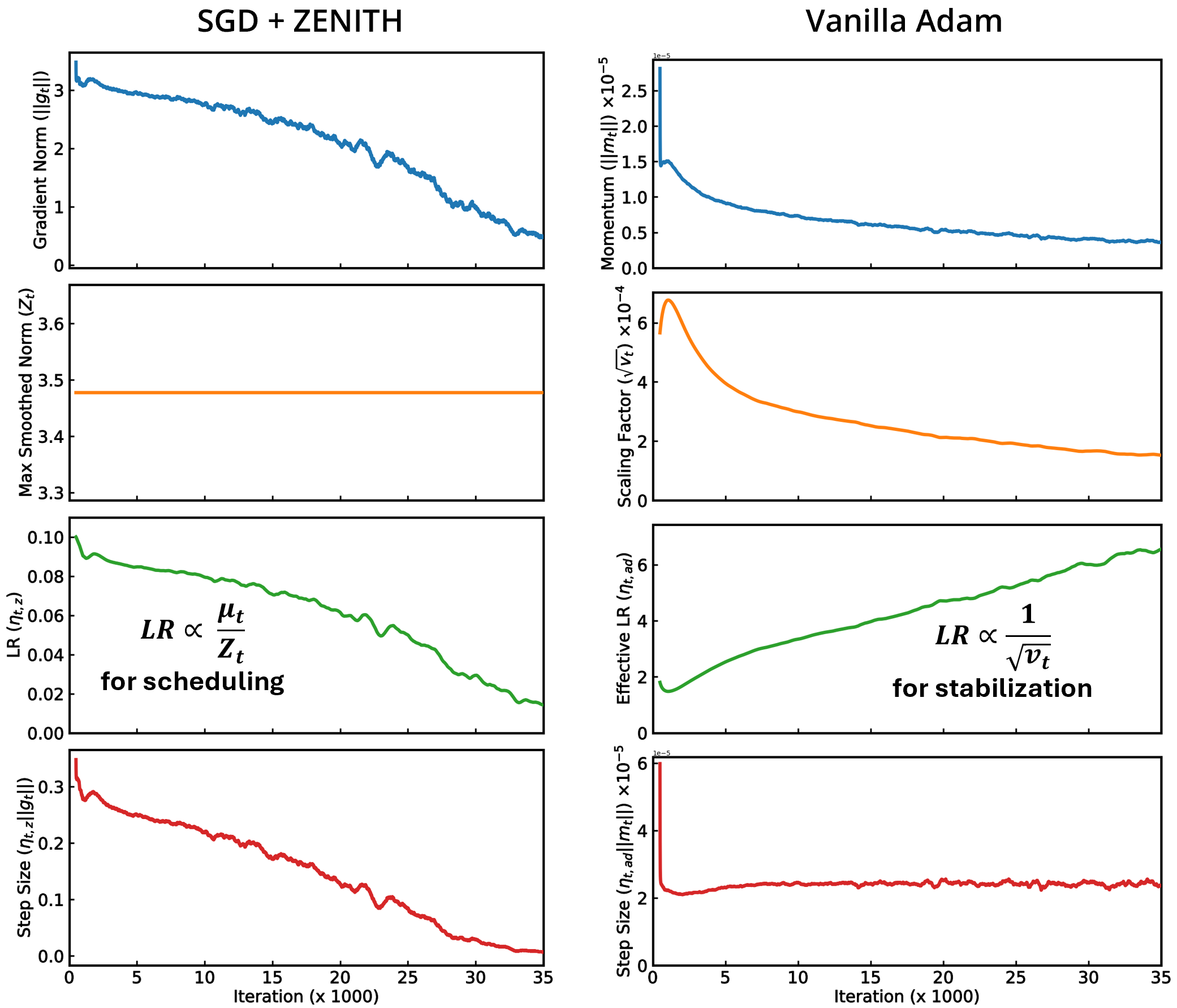}
    \captionof{figure}{Evolution of quantities comparing SGD + ZENITH with Vanilla Adam. Corresponding quantities share the same row and color. Image classification experiment B on CIFAR-10 is used here.}
    \label{fig:distinctions}

    \vspace{1em} 
    \end{center}
}] 


\textbf{Update rule of ZENITH with SGD base optimizer:}

Let $\mathbf{g}_t = \nabla\mathcal{L}(\theta_{t-1})$ denote the gradient vector.
\[
\theta_t = \theta_{t-1} - \left( \eta_0 \frac{\mu_t}{Z_t} \right) \mathbf{g}_t = \theta_{t-1} - \eta_{t,z} \mathbf{g}_t
\]

\textbf{Update rule of Vanilla Adam for each weight:}

\[
\theta_t = \theta_{t-1} - \left( \eta_0 \frac{1}{\sqrt{v_t} + \epsilon} \right) \hat{m}_t = \theta_{t-1} - \eta_{t,ad} \hat{m}_t
\]

The colored curves referred to in our following explanation are in Figure~\ref{fig:distinctions}. ZENITH's gradient norm $\|\mathbf{g}_t\|$ and Adam's momentum norm $\|\hat{m}_t\|$ are the blue curves. $\eta_{t,z}$ and $\eta_{t,ad}$ are the effective LRs.

\textbf{SGD + ZENITH.} The denominator of $\eta_{t,z}$ is $Z_t$, which is the maximum smoothed gradient norm observed during early training stages. During training, $Z_t$ remains a large constant (orange curve). Consequently, $\eta_{t,z}$ decreases (green curve) since $\mu_t$ decays. Therefore, there is a systematic reduction in step size (red curve), enabling smooth convergence.

\textbf{Vanilla Adam.} The denominator of $\eta_{t,ad}$ is dominated by $\sqrt{v_t}$. As gradients decrease, $\sqrt{v_t}$ decays (orange curve), which increases $\eta_{t,ad}$ (green curve). So the variation of the effective LR over time is the opposite of SGD + ZENITH. Hence, step sizes do not decay (red curve), causing unstable oscillations rather than convergence.

ZENITH is a scheduler for long-term LR decay of a base optimizer, whereas Adam is an optimizer focused on short-term step-size adjustments. As shown earlier, the long-term LR trajectories of SGD paired with ZENITH and vanilla Adam are completely opposite. ZENITH's LR has a positive relationship with gradient magnitudes, whereas vanilla Adam's LR has an inverse relationship. Consequently, Adam often requires a separate LR scheduler to converge stably. As demonstrated in the main text, adaptive optimizers like Adam, Adamax, and Adagrad benefit significantly from a scheduler like ZENITH, which aligns with the standard practice of using LR schedules alongside adaptive optimizers. While this section focuses on comparing SGD + ZENITH to Vanilla Adam for brevity, the same type of analysis applies to other adaptive optimizers like AdamW and Adamax. These other adaptive optimizers function similarly to Adam and display similar trends as Adam in effective LR and step size throughout training.

\twocolumn[{%
    \begin{center}
    
    \section{G. Detection and Segmentation Experiments}
    \label{sec:appendix_det_seg} 

    \vspace{1em} 
    
    \includegraphics[width=\textwidth]{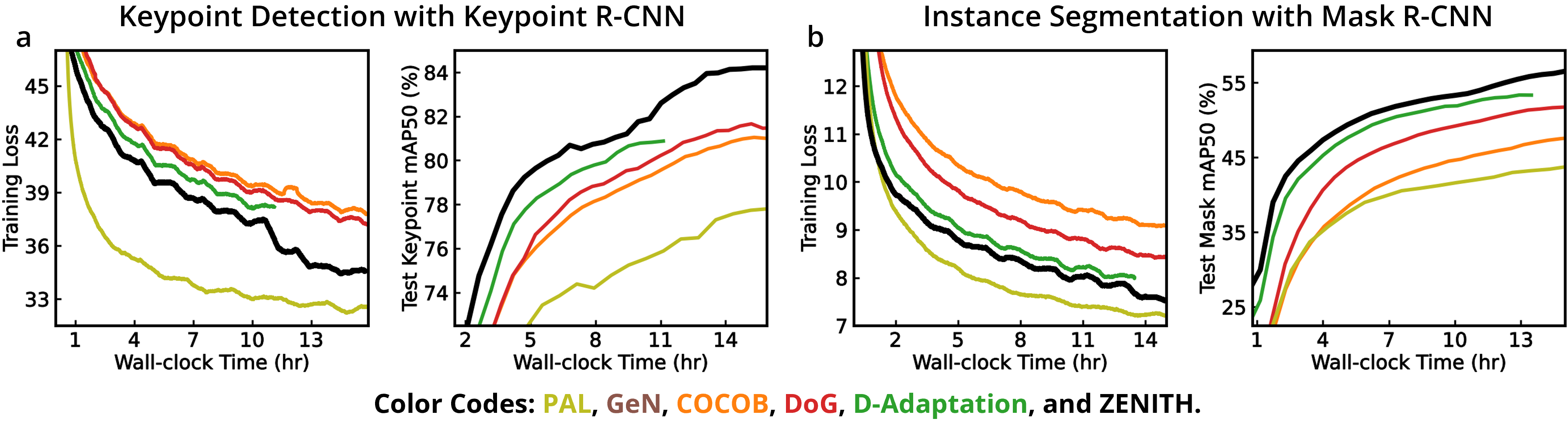}
    \captionof{figure}{Training loss and test mAP$_{50}$ curves against wall-clock time for detection and segmentation experiments. Training loss magnitudes in this figure are scaled by a factor of 10 or 100 for visual clarity. This linear transformation preserves the original curve profiles without any distortions. The loss function was not scaled during the actual training process. Curves are also smoothed using a simple moving average for visual clarity.}
    \label{fig:det_seg_plot}

    \vspace{1em} 


    \setlength{\aboverulesep}{0pt}
    \setlength{\belowrulesep}{0pt}
    \renewcommand{\arraystretch}{1.15}

    \begin{NiceTabularX}{\textwidth}{l | Y Y Y | Y Y Y}
        \CodeBefore
            \rowcolors{3}{lightgray}{white}
        \Body
        
        \hline 
        
        \Block{2-1}{Method} & 
        \multicolumn{3}{c|}{\makecell{Keypoint Detection}} & 
        \multicolumn{3}{c}{\makecell{Instance Segmentation}} \\
        
        \cline{2-4} \cline{5-7} 
        
         & \makecell{Box\\mAP$_{50}$ (\%)} & \makecell{Keypoint\\mAP$_{50}$ (\%)} & \makecell{Time\\(hr)} 
         & \makecell{Box\\mAP$_{50}$ (\%)} & \makecell{Mask\\mAP$_{50}$ (\%)} & \makecell{Time\\(hr)} \\
        
        \hline 
        
        PAL         & 72.1 $\pm$ 0.24 & 78.9 $\pm$ 0.36 & 14.4 $\pm$ 0.39 & 46.2 $\pm$ 0.21 & 44.1 $\pm$ 0.36 & 13.5 $\pm$ 0.44 \\
        GeN         & 62.9 $\pm$ 0.81 & 61.7 $\pm$ 0.46 & 1.43 $\pm$ 0.07 & 29.7 $\pm$ 0.69 & 27.2 $\pm$ 0.83 & 1.59 $\pm$ 0.08 \\
        \hline
        COCOB       & 79.6 $\pm$ 0.49 & 81.6 $\pm$ 0.42 & 15.8 $\pm$ 0.60 & 51.7 $\pm$ 0.48 & 47.9 $\pm$ 0.81 & 14.8 $\pm$ 0.37 \\
        \hline
        DoG         & 80.7 $\pm$ 4.93 & 82.3 $\pm$ 4.79 & 16.2 $\pm$ 0.89 & 55.1 $\pm$ 2.67 & 51.9 $\pm$ 2.88 & 14.4 $\pm$ 0.76 \\
        DAdapt      & 80.3 $\pm$ 0.31 & 81.8 $\pm$ 0.38 & 10.6 $\pm$ 0.28 & 57.4 $\pm$ 0.51 & 53.6 $\pm$ 0.50 & 12.4 $\pm$ 0.29 \\
        \hline
        \textbf{ZENITH} & \B{81.3 $\pm$ 0.14} & \B{84.2 $\pm$ 0.14} & \B{13.2 $\pm$ 0.25} & \B{59.3 $\pm$ 0.19} & \B{56.0 $\pm$ 0.15} & \B{14.5 $\pm$ 0.22} \\
        
        \hline 
    \end{NiceTabularX}
    \captionof{table}{Comparison of mAP$_{50}$ across schedulers and the requisite wall-clock time to achieve it. Best instantaneous combinations of mAP$_{50}$ and requisite time to achieve that mAP$_{50}$ are in bold.}
    \label{tab:optimizer_comparison_det_seg}
    \vspace{1em} 
    \end{center}

}] 

\textbf{Experimental Setup.} Experiments were conducted on the MS COCO dataset \cite{lin2014microsoft}, evaluating keypoint detection using Keypoint R-CNN and instance segmentation using Mask R-CNN \cite{he2017mask}. The ResNet-50 backbone was initialized with ImageNet-pretrained weights and was not frozen. Other architectural components were randomly initialized. Models were trained with a batch size of 16 on an NVIDIA A100 (80 GB) GPU using five data loading workers. The loss function aggregated classification, localization, and mask/keypoint-specific loss components. Test performance was evaluated three times per epoch (at iterations 2500, 5000, and 7393) using Keypoint mAP$_{50}$ and Box mAP$_{50}$ for keypoint detection, and Mask mAP$_{50}$ and Box mAP$_{50}$ for instance segmentation. The combined duration for both training and periodic testing was 24 hours. Within this time, the training phase for Keypoint R-CNN and Mask R-CNN was approximately 17 hours and 15 hours, respectively. The remaining time was for testing. Cumulative wall-clock time was recorded exclusively for the training phase. $L_2$ regularization was not used. The base optimizer was SGD.

\textbf{Experimental Results.} Several baseline schedulers were not suited for keypoint detection and instance segmentation. Some methods, such as LQA, SPS, and ALIG, produced significant overestimates of the LR, defaulting to the maximum LR cap throughout training. Others, such as Prodigy, reverted to constant LRs as well, while L4 and DoWG failed to learn at all. Consequently, only five baselines yielded valid results, which are detailed in Figure~\ref{fig:det_seg_plot} and Table~\ref{tab:optimizer_comparison_det_seg}. Figure~\ref{fig:det_seg_plot} shows that ZENITH achieves higher mAP$_{50}$ than the baselines at every wall-clock time throughout training. This indicates that ZENITH not only reaches the best final mAP$_{50}$ scores in Table~\ref{tab:optimizer_comparison_det_seg}, but also provides stronger detection and segmentation performance at any point in the training process. To be specific, Figure~\ref{fig:det_seg_plot} shows that the ZENITH mAP$_{50}$ curves are always higher than baselines', meaning it achieves each mAP$_{50}$ in less time than baselines. Interestingly, while PAL decreased the training loss most rapidly, its generalization was poor. This is attributed to the higher sharpness of its minima, and PAL's probing step size hyperparameter may require further tuning for these tasks to converge to flatter minima. The training curves for D-Adaptation are incomplete because divergent gradients ended training prematurely. Its monotonically increasing LR could have contributed to this divergence.

\twocolumn[{%
    \begin{center}
    
    \section{H. Tabular Modality Experiments}
    \label{sec:appendix_tabular} 

    \vspace{1em} 
    
    \includegraphics[width=\textwidth]{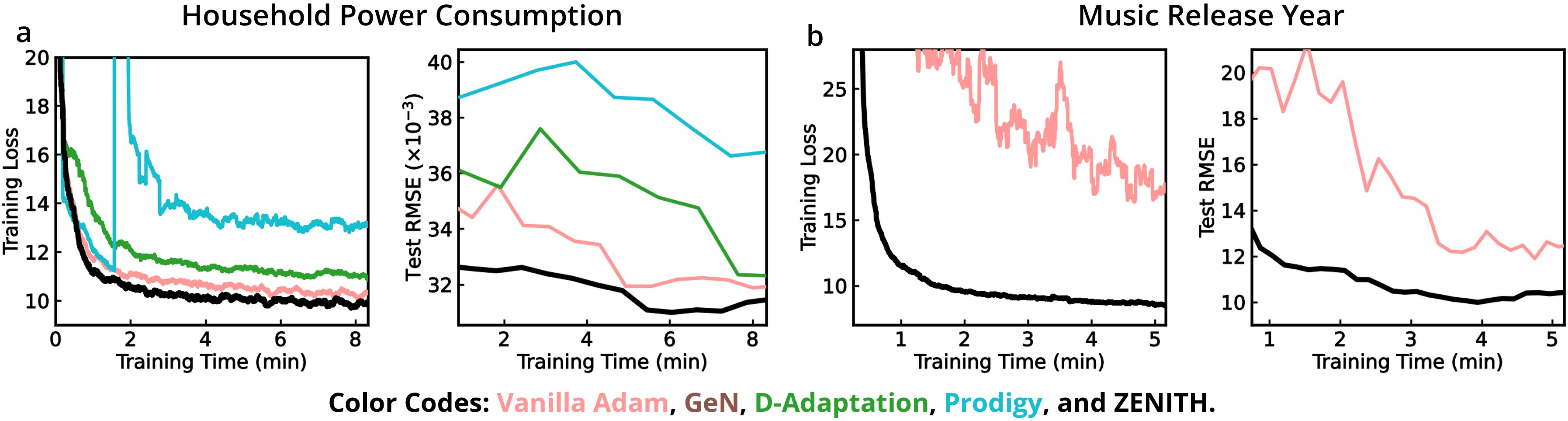}
    \captionof{figure}{Training loss and test RMSE curves against wall-clock time for the tabular modality experiments (Household Power Consumption and Music Release Year). Training loss magnitudes in this figure are scaled by a multiple of 10 for visual clarity. This linear transformation preserves the original curve profiles without any distortions. The loss function was not scaled during the actual training process. Curves are smoothed using a simple moving average for visual clarity.}
    \label{fig:tabular_plot}

    \vspace{1em} 


    \setlength{\aboverulesep}{0pt}
    \setlength{\belowrulesep}{0pt}
    \renewcommand{\arraystretch}{1.15}

    \begin{NiceTabularX}{\textwidth}{l | Y Y | Y Y}
        \CodeBefore
            \rowcolors{3}{lightgray}{white}
        \Body
        
        \hline 
        
        \Block{2-1}{Method} & 
        \multicolumn{2}{c|}{\makecell{Household Power Consumption}} & 
        \multicolumn{2}{c}{\makecell{Music Release Year}} \\
        
        \cline{2-3} \cline{4-5} 
        
         & \makecell{Test RMSE} & \makecell{Time (min)} 
         & \makecell{Test RMSE} & \makecell{Time (min)} \\
        
        \hline 
        
        Adam        & $0.03117 \pm 0.0004117$ & $5.757 \pm 2.537$ & $9.938 \pm 0.1778$ & $4.389 \pm 0.4382$ \\
        \hline
        GeN & --- & --- & --- & --- \\
        \hline
        DAdapt      & $0.03189 \pm 0.0003305$ & $6.857 \pm 1.967$ & --- & --- \\
        Prodigy     & $0.03228 \pm 0.0004405$ & $4.045 \pm 3.004$ & --- & --- \\
        \hline
        \textbf{ZENITH} & \textbf{0.03078} $\pm$ \textbf{0.0001742} & \textbf{5.840} $\pm$ \textbf{2.278} & \textbf{9.905} $\pm$ \textbf{0.3191} & \textbf{3.494} $\pm$ \textbf{1.390} \\
        
        \hline 
    \end{NiceTabularX}
    \captionof{table}{Comparison of Test RMSE and required wall-clock training time across schedulers for tabular datasets. Best combinations of RMSE and requisite time to achieve that RMSE are in bold. An em-dash (---) indicates the algorithm diverged or failed to learn.}
    \label{tab:optimizer_comparison_tabular}
    \vspace{1em} 
    \end{center}
}]

\textbf{Experimental Setup.} For the regression task on the Household Power Consumption dataset \cite{dua2019individual}, the objective was to predict the global active power using the remaining continuous variables (global reactive power, voltage, global intensity, and the three sub-metering metrics). Temporal features (date and time) and rows containing missing values were removed prior to modeling. The dataset was partitioned into an 80\% training set and a 20\% test set. All input features were normalized using standard scaling (zero mean and unit variance), with the scaler fitted on the training data. The model architecture was a Multi-Layer Perceptron (MLP) consisting of three hidden layers, each containing 512 neurons and utilizing ReLU activation functions, followed by a single-node linear output layer. A Mean Squared Error (MSE) loss function and batch size of 128 were used. Model performance was evaluated using Root Mean Squared Error (RMSE). We tracked the wall-clock training time, excluding the testing overhead. Three runs were conducted with random seeds of 42, 43, and 44. The training time was 10 minutes on an NVIDIA A100 GPU. The Adam optimizer with an initial LR of $10^{-3}$ was the base optimizer. 

For the Music Release Year dataset \cite{bertin2011million}, the objective was to predict the release year of a song based on 90 continuous audio features. The first column of the dataset was the target regression variable, with the subsequent 90 columns serving as the input features. The training time was 5 minutes. All other configurations remained identical to the power consumption experiment. 

Most baselines from the Related Work section are designed only for SGD and are incompatible with Adam. Therefore, we tested the Adam-compatible baselines: DAdaptation, Prodigy, and GeN.

\textbf{Experimental Results.}  Figure~\ref{fig:tabular_plot} and Table~\ref{tab:optimizer_comparison_tabular} shows that ZENITH achieves lower test RMSE than the baselines at every wall-clock time throughout training. This indicates that ZENITH provides stronger tabular-regression performance at any point in the training process, while Table~\ref{tab:optimizer_comparison_tabular} confirms that it reaches the lowest test RMSE on both datasets. GeN failed to learn on these tasks. As in previous experiments, its LR decayed too quickly, and in these cases, the decay was so fast that it prevented any learning. Additionally, DAdaptation and Prodigy diverged on the Music Release Year dataset because their LR growth was too aggressive.


\end{document}